\documentclass[acmtog,screen]{acmart}

\acmJournal{TOG}

\usepackage{overpic}
\usepackage{arydshln}
\usepackage{amsmath}
\usepackage{graphics}
\usepackage{placeins} %?
\usepackage{makecell} %?
\usepackage{wrapfig}
\usepackage{amssymb}
\usepackage{multirow}
\usepackage{enumerate}
\usepackage{color}

\newcommand{\FF}[1]{{#1}}
\newcommand{\FFF}[1]{{#1}}

\usepackage{xspace}
\newcommand{\name}{\textsl{SATO}\xspace}
\usepackage{algorithm}
\usepackage{algpseudocode}  % 注意是这个包，不是 algorithmic
\usepackage{enumitem}

\setcopyright{cc}
\setcctype{by-nc-nd}
\acmJournal{TOG}
\acmYear{2026} \acmVolume{45} \acmNumber{4} \acmArticle{75}
\acmMonth{7} \acmDOI{10.1145/3811286}

\begin{document}

\title{Strips as Tokens: Artist Mesh Generation with Native UV Segmentation}
% \title{SATO: Strips as Tokens for Autoregressive Artist Mesh Generation}

% \affiliation{%
\author{Rui Xu}
\authornotemark[1]
\affiliation{\institution{The University of Hong Kong, Deemos Technology Co., Ltd.} 
\authornote{Equal contribution. ‡ Project lead. † Corresponding authors.}
\country{China}}
\email{ruixu1999@connect.hku.hk}

\author{Dafei Qin}
\authornotemark[1]
\affiliation{\institution{The University of Hong Kong, Deemos Technology Co., Ltd.} 
\country{China}}
\email{qindafei@connect.hku.hk}

\author{Kaichun Qiao}
\affiliation{\institution{ShanghaiTech University, Deemos Technology Co., Ltd.} 
\country{China}}
\email{qiaokch2022@shanghaitech.edu.cn}

\author{Qiujie Dong}
\affiliation{\institution{Shandong University} 
\country{China}}
\email{qiujie.jay.dong@gmail.com}

\author{Huaijin Pi}
\affiliation{  \institution{The University of Hong Kong}
\country{China}}
\email{huaijinpi@connect.hku.hk}

\author{Qixuan Zhang}
\authornotemark[3]
\affiliation{\institution{ShanghaiTech University, Deemos Technology Co., Ltd.} 
\country{China}}
\email{zhangqx1@shanghaitech.edu.cn}

\author{Longwen Zhang}
\affiliation{\institution{ShanghaiTech University, Deemos Technology Co., Ltd.} 
\country{China}}
\email{zhanglw2@shanghaitech.edu.cn}

\author{Lan Xu}
\authornotemark[2]
\affiliation{\institution{ShanghaiTech University} 
\country{China}}
\email{xulan1@shanghaitech.edu.cn}

\author{Jingyi Yu}
\affiliation{\institution{ShanghaiTech University} 
\country{China}}
\email{yujingyi@shanghaitech.edu.cn}

\author{Wenping Wang}
\affiliation{  \institution{Texas A\&M University}
\country{USA}}\email{wenping@tamu.edu}

\author{Taku Komura} 
\authornotemark[2]
\affiliation{  \institution{The University of Hong Kong}
\country{China}}
\email{taku@cs.hku.hk}


\begin{abstract}
Recent advancements in autoregressive transformers have demonstrated remarkable potential for generating artist-quality meshes.
However, the token ordering strategies employed by existing methods typically fail to meet professional artist standards, where coordinate-based sorting yields inefficiently long sequences, and patch-based heuristics disrupt the continuous edge flow and structural regularity essential for high-quality modeling.
To address these limitations, we propose Strips as Tokens (\name), a novel framework with a token ordering strategy inspired by triangle strips.
By constructing the sequence as a connected chain of faces that explicitly encodes UV boundaries, our method naturally preserves the organized edge flow and semantic layout characteristic of artist-created meshes.
A key advantage of this formulation is its unified representation, enabling the same token sequence to be decoded into either a triangle or quadrilateral mesh.
This flexibility facilitates joint training on both data types: large-scale triangle data provides fundamental structural priors, while high-quality quad data enhances the geometric regularity of the outputs. 
Extensive experiments demonstrate that \name consistently outperforms prior methods in terms of geometric quality, structural coherence, and UV segmentation. 
\end{abstract}




\begin{teaserfigure}
  \centering
  % \vspace{-2mm}
  \includegraphics[width=\textwidth]{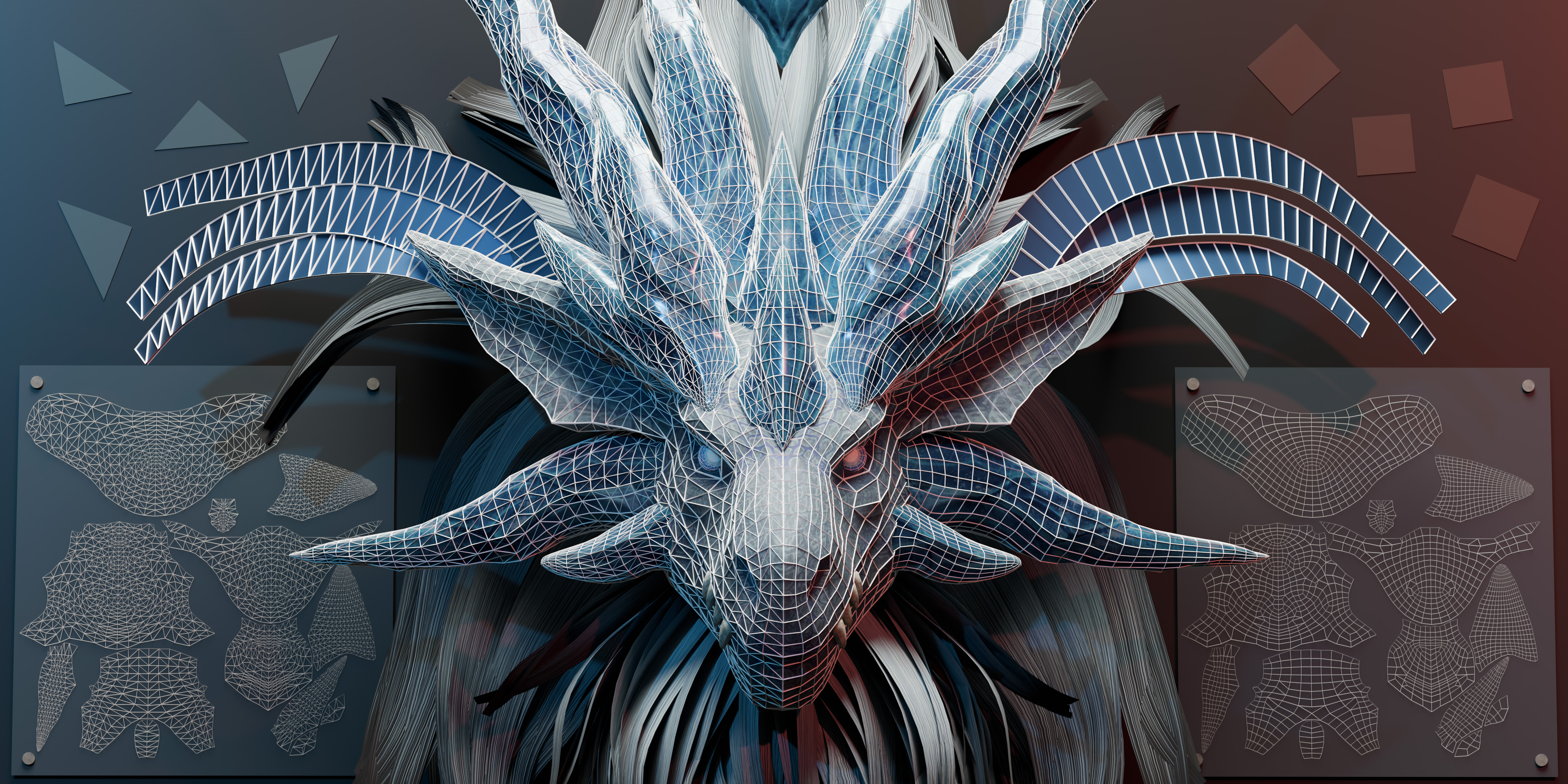}
  \vspace{-1mm}
  % \makebox[\linewidth][l]{ \textbf{(a)~Initial normal vectors \hspace{1.9cm} (b)~20 iterations \hspace{2.5cm} (c)~40 iterations   \hspace{2.1cm} (d)~Reconstruction}}
  \vspace{-7mm}
  \caption{
  \textbf{Strips as Tokens (SATO) enables unified, high-quality artist mesh generation with native UV segmentation.}
Our strip-based tokenizer supports both triangle (left) and quad (right) meshes without retraining and automatically segments UV charts (side) during autoregressive generation. 
  } 
  \label{fig:teaser}
\end{teaserfigure}


\maketitle
\newcommand{\rspace}{\mathbb{R}}
\newcommand{\vorcell}{\Omega^{vor}}
\newcommand{\anypoint}{\mathbf{x}}
\newcommand{\sample}{\mathbf{p}}
\newcommand{\query}{\mathbf{q}}
\newcommand{\cross}{\mathcal{A}}
\newcommand{\area}{a}
\newcommand{\normal}{\mathbf{n}}
\newcommand{\numV}{M}
\newcommand{\numP}{N}
\newcommand{\ie}{\textit{i.e., }}
\newcommand{\eg}{\textit{e.g., }}

\newcommand{\phj}[1]{{\textcolor{purple}{[PHJ: #1]}}}

\section{Introduction}
Artist-created meshes remain the dominant representation for 3D assets: they facilitate direct surface editing, provide precise control over connectivity and edge flow, and form the backbone of downstream stages such as deformation, simulation, and texture mapping. In contrast to meshes produced by generic remeshing algorithms, artist meshes usually adhere to consistent structural conventions. For instance, they often favor right-angled triangles over equilateral ones; triangles tend to align anisotropically with principal and secondary curvature directions; and sampling density increases in high-curvature regions while remaining sparse on flatter areas. These conventions profoundly impact rigging and animation quality, texturing workflows, and the long-term maintainability of production assets.

\begin{figure}[!tp]
    \centering
    \begin{overpic}[width=\linewidth]{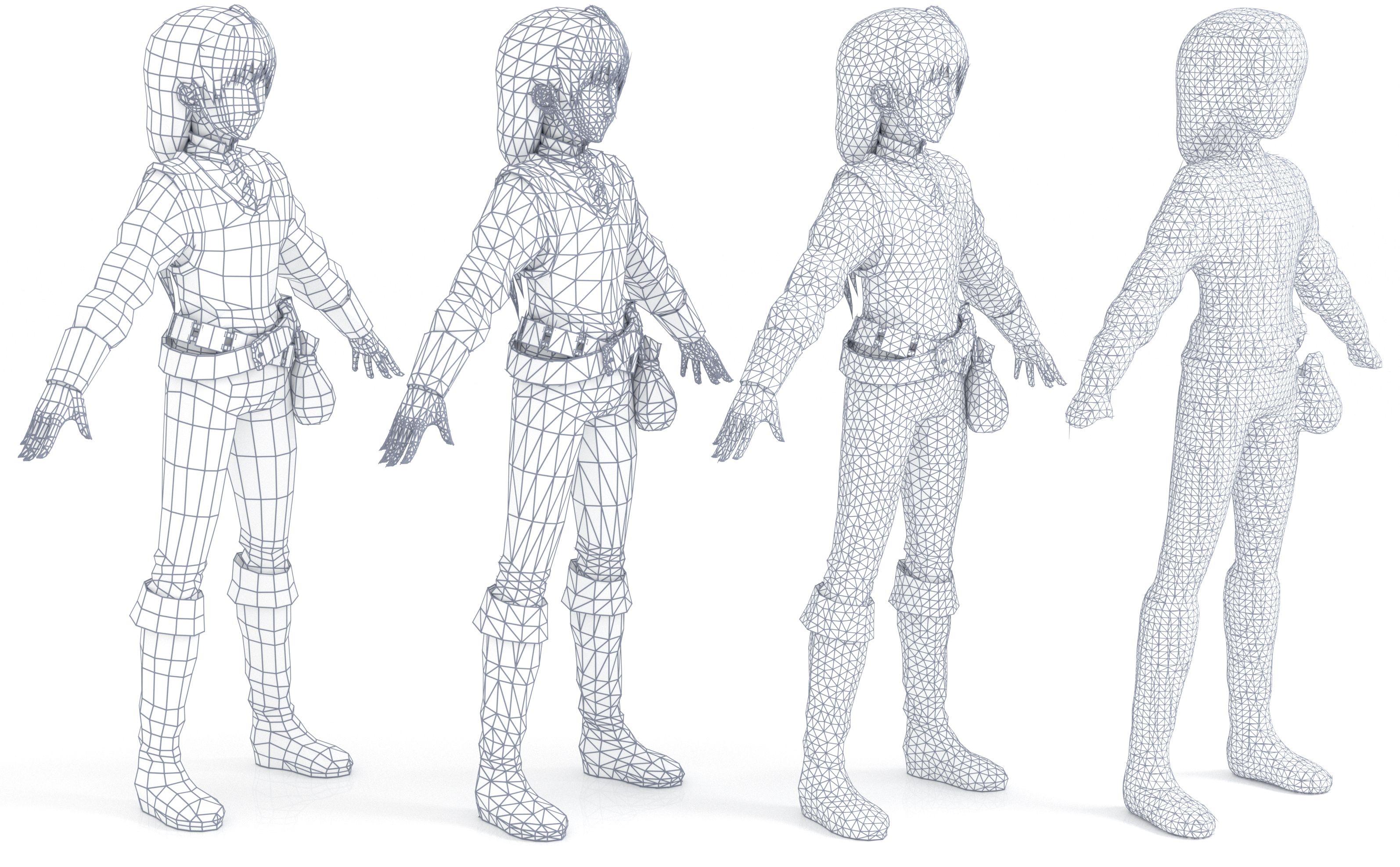}
    \put(7,  -2){\textbf{Artist Quad}}
    \put(33, -2){\textbf{Artist Tri}}
    \put(54, -2){\textbf{Delaunay}}
    \put(73, -2){\textbf{Marching Cubes}}
    \end{overpic}
    \vspace{-5mm}
    \caption{\textbf{Artist meshes differ markedly from geometry-processed ones.} Here we show quadrilateral and triangular meshes constructed by artists, as well as meshes created using geometric processing methods (such as Delaunay-style remeshing~\cite{xu2024cwf} and Marching Cubes~\cite{MarchingCubes}).}
    \label{fig:artist}
    \vspace{-4mm}
\end{figure}

Generating high-quality 3D meshes that meet professional production standards is particularly challenging: a mesh must not only capture accurate high-fidelity geometry but also possess regular topology (i.e., clean edge flow) and semantic layouts (i.e., UV mapping) to be compatible with animation and rendering pipelines. Recently, autoregressive modeling has emerged as a promising alternative, treating mesh generation as a sequence prediction task~\citep{Meshgpt,chen2024meshxl,chen2024meshanythingv2,hao2024meshtron}. By learning a distribution over discrete tokens, these methods attempt to capture geometric patterns directly. Early approaches typically rely on coordinate-based ordering~\citep{hao2024meshtron,chen2024meshxl}. These methods directly tokenize the mesh by converting vertex coordinates into sorted triplets, each defining a tuple of quantized 3D coordinates. However, this fine-grained representation results in excessively long sequences. To address this, more recent methods~\citep{bpt,zhao2025deepmesh,xu2025meshmosaic} employ patch-based tokenization that relies on Delaunay-style heuristics to organize the token order, thereby significantly shortening the sequence length. However, this approach inherently sacrifices the continuous surface curvature direction and coherent edge flow central to artist-created meshes, as Delaunay-style triangulation prioritizes mathematical compactness (e.g., maximizing minimum angles) over structural regularity. Fig.~\ref{fig:artist} highlights the distinct contrast between artist meshes, Delaunay-style meshes~\cite{xu2024cwf}, and meshes obtained by Marching Cubes~\cite{MarchingCubes}.

\begin{wrapfigure}{r}{3.5cm}
\vspace{-1.5mm}
  \hspace*{-4mm}
  \centerline{
  \includegraphics[width=45mm]{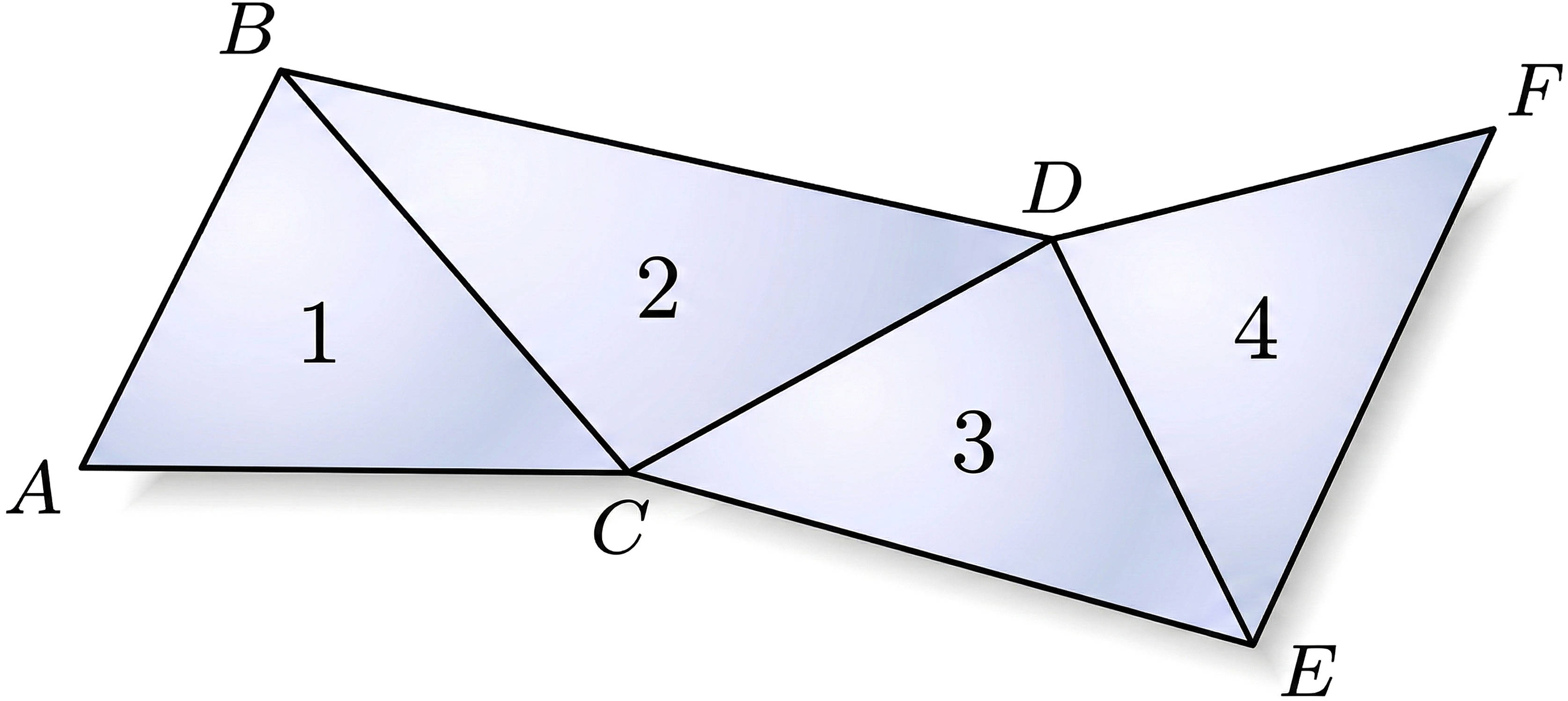}
  }
  \vspace*{-6mm}
\end{wrapfigure}
Our key insight stems from the \textbf{triangle strip}, a classic concept representing a sequence of triangles that share vertices, thereby offering a memory-efficient mesh storage format. Each newly appended vertex deterministically forms a new triangle with the two preceding vertices, yielding a compact encoding that inherently couples connectivity with local surface continuity and directly exposes the flow-like structure of the original mesh. These properties create a strong structural alignment with artist-created topology, motivating the incorporation of the strip formulation into our tokenization strategy.

% Our key insight draws from \textbf{triangle strip}, which is a classic concept that represents a sequence of triangles in a mesh that share vertices, offering a more memory-efficient way to store meshes. 
% Each newly appended vertex deterministically forms a new triangle with the two preceding vertices, yielding a compact encoding that couples connectivity with local surface continuity and directly exposes flow-like structure in the original mesh. 
% These properties create a strong structural alignment with artist-created topology, motivating us to incorporate the strip formulation into the tokenization.
In this paper, we propose a novel framework, named \textbf{Strips as Tokens (\name)}, for generating artist-quality 3D meshes. Our core innovation lies in a strip-based tokenization strategy that organizes vertex ordering according to the topological definition of strips. Specifically, we construct the sequence as a connected chain of faces where each consecutive pair shares a common edge, a property that inherently aligns with the organized edge flow of artist meshes. Crucially, a key advantage of this formulation is that the unified vertex ordering enables a dual interpretation. Leveraging the topological fact that a quadrilateral naturally decomposes into two adjacent triangles, our framework allows the same token sequence to be decoded as either a triangle or a quadrilateral mesh. This flexibility facilitates the synergistic use of both data types: extensive triangle data provides fundamental structural priors, while high-quality quad data further enhances the geometric regularity of triangle outputs. Furthermore, we support native UV segmentation by extending the token vocabulary with specialized segmentation tokens. This mechanism encodes UV island boundaries directly into the token sequence without sacrificing compression efficiency, enabling the model to explicitly predict semantic partitioning.
% In this paper, we propose a novel framework, named \textbf{Strips as Tokens (\name)}, for artist 3D mesh generation.
% Our innovation is in a strip-based tokenization strategy that organizes the vertex ordering based on the topological definition of strips.
% Specifically, we construct the sequence as a connected chain of faces where each consecutive pair shares a common edge, a property that aligns perfectly with the organized edge-flow of artist meshes.
% Additionally, a key advantage of this formulation is that the unified vertex ordering enables a dual interpretation.
% Leveraging the topological fact that a quadrilateral naturally decomposes into two adjacent triangles, our framework allows the same token sequence to be decoded as either a triangle or a quadrilateral mesh.
% This flexibility enables the joint use of both data types: the extensive triangle data provides fundamental structural priors while the high-quality quad data further improves the quality of triangle outputs.
% Furthermore, we support native UV segmentation by extending the token vocabulary with specialized segmentation tokens. 
% This mechanism encodes UV island boundaries directly into the token sequence without sacrificing compression efficiency, enabling the model to explicitly predict the semantic partitioning.

We evaluate \name across diverse datasets and tasks, observing consistent improvements over prior methods in geometric fidelity, structural coherence, and UV-aware generation. These results highlight the critical role of representation design in autoregressive mesh generation, suggesting that artist-aligned tokenization is a key ingredient for making such models both learnable and practical.
In summary, we make the following contributions:
\begin{itemize}[leftmargin=2em, itemsep=0.2em] 
\item \textbf{Strip tokenization.} We propose an artist-aligned strip-based serialization that preserves edge-flow coherence, achieves high compression efficiency, and makes the sequence structure easier for the model to learn.
\item \textbf{Unified tri/quad decoding.} A single token sequence supports both triangle and quad decoding, enabling triangle and quad data to synergistically reinforce each other through fine-tuning and bidirectional prior transfer. 
\item \textbf{Native UV segmentation.} We explicitly encode UV island boundaries with dedicated tokens, making \name the first autoregressive framework to simultaneously generate mesh geometry and UV chart partitions. \end{itemize}

\section{Related Work}

\subsection{3D Generation} A growing body of work synthesizes 3D assets utilizing implicit or hybrid representations, including signed distance fields, occupancy fields, and multi-view neural pipelines. Representative systems such as Wonder3D~\citep{long2024wonder3d}, TRELLIS~\citep{TRELLIS}, TRELLIS.2~\citep{TRELLIS2}, CLAY~\citep{CLAY}, and Hunyuan3D-2.5~\citep{lai2025hunyuan3d} achieve impressive end-to-end generation of textured geometry, often conditioned on text or images. More recently, several methods have shifted towards integrating stronger structural priors aimed at editability and decomposition. CraftsMan3D~\citep{li2024craftsman3d} moves toward mesh-native outputs via 3D diffusion augmented by an (interactive) geometry refiner. OmniPart~\citep{OmniPart2025} and Ultra3D~\citep{Ultra3D2025} emphasize part-aware synthesis through semantic decoupling and part attention. BANG~\citep{bang} explores generative “exploded” dynamics for controllable asset division, and CAST~\citep{cast} targets component-aligned reconstruction for multi-object scenes from a single image. Despite these advances, the final geometry often necessitates conversion to explicit meshes via iso-surfacing (commonly Marching Cubes~\citep{MarchingCubes}) or related extraction procedures, which typically results in dense triangle meshes with connectivity that is poorly aligned with authoring conventions. Consequently, despite high visual fidelity, substantial post-processing is still required to obtain compact, editable, production-friendly meshes. Reliably producing truly artist-friendly meshes—clean topology with oriented regularities—therefore remains an open challenge, motivating our focus on artist mesh generation.

\subsection{Mesh Generation} \subsubsection{Triangle Mesh} Autoregressive mesh generation has emerged as a compelling paradigm for producing compact, artist-like triangle meshes by predicting discrete symbols in a causal order. MeshGPT~\citep{Meshgpt} is an early representative that learns a discrete vocabulary and generates meshes as sequences, demonstrating that transformer-style decoding can yield sharp yet compact triangulations. Subsequent work has expanded fidelity and scale by refining tokenization and decoding strategies. MeshAnything~\citep{chen2024meshanything} and MeshAnythingV2~\citep{chen2024meshanythingv2} propose adjacency-aware tokenizations to shorten sequences and improve controllability, while MeshXL~\citep{chen2024meshxl} explores coordinate-field-style representations for large-scale sequential modeling. EdgeRunner~\citep{tang2024edgerunner} further improves token efficiency with classical-mesh-inspired serialization, and introduces an autoregressive auto-encoder that maps variable-length meshes into compact latent codes. Concurrently, network-oriented efforts address the computational bottlenecks of long-context decoding. Meshtron~\citep{hao2024meshtron} scales triangle mesh generation to substantially higher face counts via an hourglass design with sliding-window inference, and iFlame~\citep{wang2025iflame} interleaves full and linear attention to reduce cost while preserving quality.

% Compression-oriented strategies offer another dimension of progress. 
% Complementing these scaling efforts, another stream of work prioritizes sequence compression.
Beyond architectural innovations, recent approaches have significantly improved performance through sequence compression, distributed processing, and optimized serialization strategies.
BPT~\citep{bpt} reduces context length via blocked and patchified representations, enabling higher-resolution geometry under limited sequence budgets, and DeepMesh~\citep{zhao2025deepmesh} extends such compressed representations with preference optimization to better match human judgments. TreeMeshGPT~\citep{lionar2025treemeshgpt} proposes a dynamic tree-based sequencing scheme that adapts next-token prediction to mesh growth. Nautilus~\citep{wang2025nautilus} studies locality-aware encoding and decoding to better preserve local manifold structure under compression. FastMesh~\citep{fastmesh} further decouples geometry and connectivity by generating vertices autoregressively and then predicting adjacency in parallel with a bidirectional transformer, enabling substantially faster artistic mesh synthesis. MeshRipple~\citep{lin2025meshripple} expands meshes from a dynamically maintained frontier using topology-aligned BFS tokenization and a global memory mechanism, improving structural completeness by retaining long-range topological context. Mesh-RFT~\citep{MeshRFT} targets post-training quality via face-level masked preference optimization with topology-aware scoring, enabling localized corrections while maintaining global coherence. MeshMosaic~\citep{xu2025meshmosaic} increases the number of triangle faces by adopting a part-based, local-to-global processing strategy with explicit interaction awareness across parts. MeshSilksong~\citep{song2025meshSilksong} adopts a weaving-style serialization that visits each vertex only once, substantially shortening sequences while promoting manifold, watertight meshes with consistent normals.

Collectively, these works improve scalability and structural fidelity through advances in sequencing, decoding, and post-training objectives. Despite these advances, most tokenizations remain fundamentally \emph{triangle-centric}, in that they rely on individual triangles, or their immediate adjacency, as the primary unit of generation. Higher-order organization, continuous surface runs, stable edge flow, and coherent region growth must then emerge implicitly from many local triangle-level decisions. This misalignment hinders the faithful capture of mid-level regularities that artists intentionally embed in production meshes. 
\name bridges this gap by elevating triangle strips to the token level, providing a compact primitive that couples connectivity with local continuity and encourages coherent surface progression.
\FF{It is worth noting that triangle strip decomposition has a long history in classical computer graphics, where various heuristic and graph-based stripification algorithms have been developed for efficient rendering of both triangle~\cite{xiang1999stripification,porcu2003iterative,vanvecek2007comparison} and quadrilateral~\cite{vanecek2005quadstrip} meshes.}

% However, MeshSilksong achieves compression through layered traversal with compact layer-adjacency encoding, whereas \name, besides generating an artist-aligned triangle-strip primitive that preserves edge flow by construction, also supports unified tri and quad decoding and native UV-boundary tokens.

% Collectively, these lines of work improve scalability and structural fidelity through better sequencing, decoding, and post-training objectives; in contrast, \name focuses on \emph{artist-aligned} primitives that directly expose edge-flow as the modeling unit.
% \XR{todo: add MeshSilk FastMesh MeshRipple  Mesh-RFT}
% \XR{clear diff with meshsilk}

\subsubsection{Quad Mesh} Compared to triangle meshes, quad-dominant meshes are often favored in production due to their regular edge flow and favorable deformation behavior. However, generating quads directly presents significant challenges, as it necessitates maintaining higher-order consistency beyond local triangulation decisions. A common strategy is therefore to first synthesize a triangle mesh and then promote quad-compatibility through scoring or post-processing. Mesh-RFT~\citep{MeshRFT} encourages quad-friendly topology via preference optimization with topology-aware rewards computed after tri-to-quad merging. Conversely, QuadGPT~\citep{QuadGPT} targets quad dominance more directly by natively modeling mixed triangle and quad faces within a unified sequence, subsequently refining topology through topology-aware post-training.

Beyond quad-dominant meshes, pure-quad meshes impose even stricter regularity requirements, particularly for edge-aligned flow and globally consistent structure. NeurCross~\citep{Dong2025NeurCross} introduces a proxy surface to implicitly align quad edge directions with principal curvature directions, but it remains computationally expensive. CrossGen~\citep{Dong2025CrossGen} improves efficiency and generalization by training a VAE to enable fast synthesis of high-quality pure-quad meshes. Nevertheless, these pipelines still depend heavily on a well-structured cross field as an explicit guiding signal, which makes fully end-to-end generation of production-quality pure-quad meshes from raw inputs difficult. In contrast, \name circumvents this dependency by directly modeling strip-consistent edge flow, enabling the one-step generation of high-quality quad meshes.
\begin{figure*}[!tp]
    \centering
    \vspace{3mm}
    \begin{overpic}[width=\linewidth]{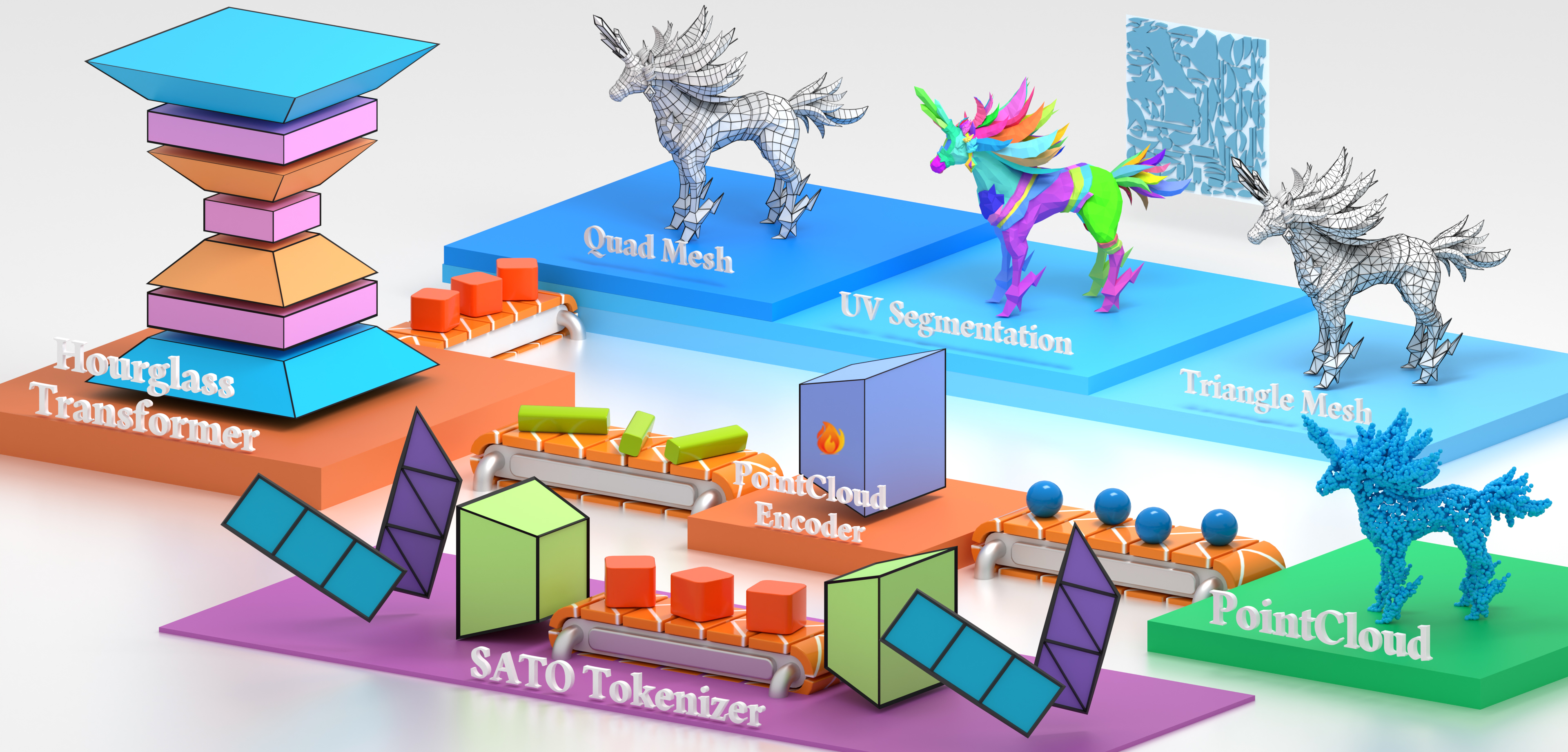}
    \end{overpic}
    \vspace{-8mm}
    \caption{\textbf{The Pipeline of \name.} \name uses a strip-based tokenizer to encode/decode both triangle and quad meshes as a unified discrete sequence. Conditioned on an input point cloud, a learnable point-cloud encoder cross-attends to the core Hourglass Transformer, which autoregressively generates token sequences that are decoded into triangle or quad meshes with native UV segmentation.}
    \label{fig:ppl}
    \vspace{-4mm}
\end{figure*}

\subsection{UV Segmentation} Production-ready meshes must support not only geometry and connectivity, but also efficient texturing workflows. Many systems in the broader 3D generation pipeline output textured assets, such as Wonder3D~\citep{long2024wonder3d}, CLAY~\citep{CLAY}, and Hunyuan3D-2.5~\citep{lai2025hunyuan3d}; however, UV unwrapping and seam placement are typically relegated to downstream stages and handled by separate parameterization and atlasing tools, rather than being integrated as constraints maintained during mesh synthesis. Similarly, most autoregressive mesh generators prioritize geometry and topology, and either omit UVs entirely or re-segment charts after generation. This decoupling disrupts artist-style seam structure and adds nontrivial post-processing overhead. In contrast, \name incorporates UV segmentation as an intrinsic part of the generative representation, by organizing strip sequences within UV regions and inserting explicit region delimiters, so that UV boundaries can be preserved and recovered during generation.

Recent learning-based methods improve UV unwrapping by explicitly learning seam placement. SeamGPT~\citep{li2025auto} and ArtUV~\citep{artuv} follow a production-inspired pipeline, where a GPT-based seam predictor proposes semantically meaningful cuts and a learned module refines an initialized UV map. However, these approaches remain multi-stage and initialization-dependent, and often lack explicit optimization for global packing efficiency. Nuvo~\citep{nuvo} models UVs as a neural field and optimizes them over visible surface points, which reduces fragmentation on challenging geometry. FAM~\citep{fam} similarly learns global free-boundary parameterization directly on surface points in an unsupervised manner, reducing reliance on high-quality meshes. PartUV~\citep{PartUV} leverages semantic part decomposition to reduce chart fragmentation under a distortion budget, coupling charting with parameterization and packing. While robust on generated meshes, it depends on the quality of part segmentation and introduces additional stages, which compromises the end-to-end nature of the pipeline and can become unstable when parts are ambiguous.

\section{Preliminaries}

\subsection{Triangle Strips}
\label{sec:pre_TriangleStrips}
A triangle strip~\cite{isenburg2001trianglestrip} is a compact encoding of a connected sequence of triangles where adjacent triangles share an edge.
Instead of storing triangles independently (as a triangle list), a strip represents triangles by an ordered vertex sequence
$\mathcal{S}=(v_1, v_2, \dots, v_m)$, which implicitly defines $m-2$ triangles:
\begin{equation}
    f_i = (v_i,\, v_{i+1},\, v_{i+2}), \qquad i=1,\dots,m-2.
\end{equation}
Consecutive triangles $f_i$ and $f_{i+1}$ share the edge $(v_{i+1}, v_{i+2})$, so each new triangle introduces only one new
vertex index, yielding a highly efficient representation for rendering and storage. 

\begin{wrapfigure}{r}{3.7cm}
\vspace{-4.5mm}
  \hspace*{-4mm}
  \centerline{
  \includegraphics[width=45mm]{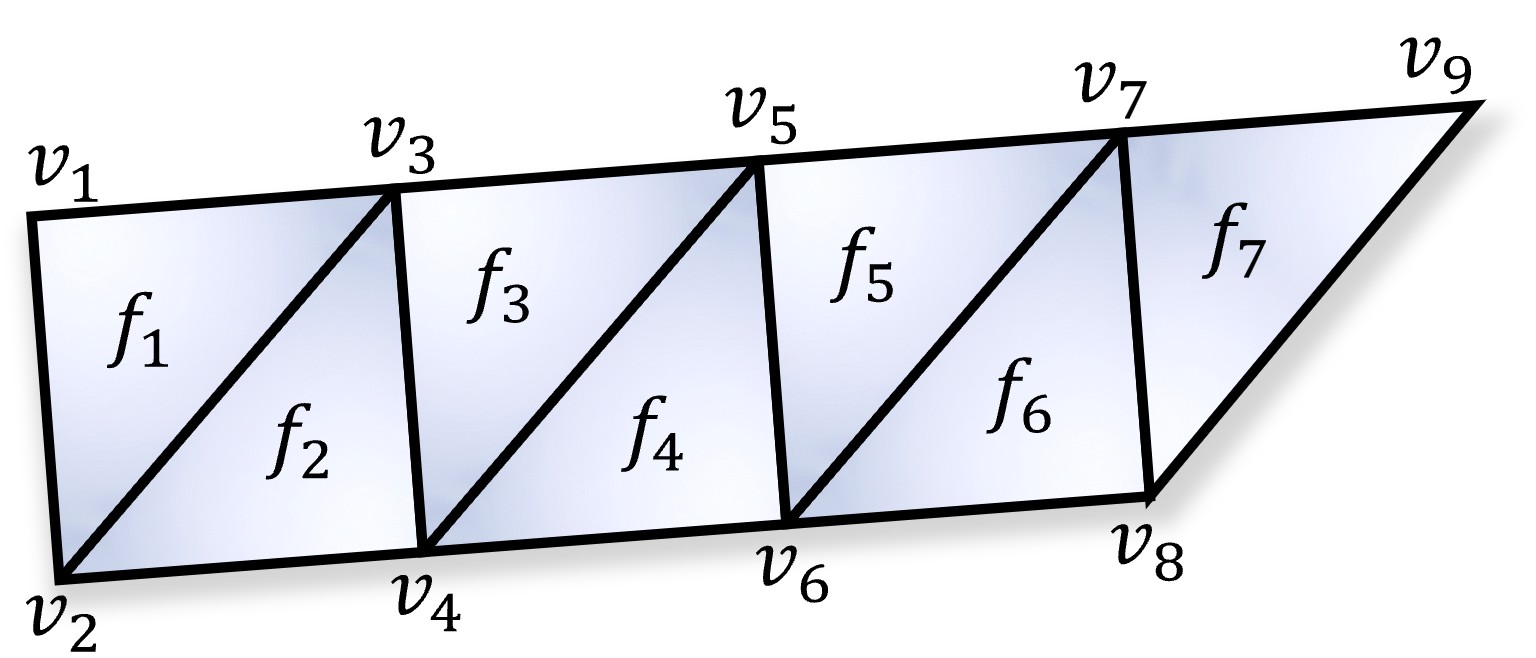}
  }
  \vspace*{-4mm}
\end{wrapfigure}
Because the vertex order alternates along the strip, the face orientation flips between neighboring triangles.
To maintain a consistent orientation (e.g., counterclockwise), one commonly reorders every other triangle as
$f_i'=(v_i, v_{i+2}, v_{i+1})$ for even $i$ (or equivalently toggles a parity flag during decoding).
In practice, a mesh can be decomposed into a set of strips.
From artists' perspective, this representation aligns well with how surfaces are commonly laid out and created:
modelers often build geometry by extending a boundary and adding faces incrementally, forming long, coherent strips of triangles with stable local connectivity.
Such strip-like structure preserves a clear sequential order and strong local adjacency, which makes it easier to capture (and later regenerate) the mesh flow and regularity of artist meshes compared to treating faces as an unordered triangle list.

\subsection{Autoregressive Mesh Generation Framework}
\label{sec:pre_ARmesh}
% Autoregressive Transformers have demonstrated strong capability for artist mesh generation in recent years. 
% Starting from MeshGPT~\cite{Meshgpt}, a series of autoregressive approaches has been proposed, including MeshAnything~\cite{chen2024meshanything,chen2024meshanythingv2}, Meshtron~\cite{hao2024meshtron}, EdgeRunner~\cite{tang2024edgerunner}, and others~\cite{bpt,zhao2025deepmesh}.
% These methods typically quantize mesh vertex coordinates to a fixed discrete resolution (e.g., 256 or 512), and then design a learnable or rule-based tokenizer to convert a triangle mesh into token sequence.
% For instance, MeshAnything~\cite{chen2024meshanything,chen2024meshanythingv2} trains a VQ-VAE to encode triangle meshes into discrete codes, whereas BPT~\cite{bpt} and DeepMesh~\cite{zhao2025deepmesh} rely on handcrafted serialization rules to impose a canonical ordering and further compress geometry via multi-level coordinate representations.

% A key practical challenge is that triangle meshes often contain thousands (or more) faces, which directly translates into extremely long token sequences and makes autoregressive training and inference expensive.
% Meshtron~\cite{hao2024meshtron} addresses this issue through a truncated sliding-window training scheme together with an hourglass Transformer, where the model is trained only on local subsequences and then performs architecture-level extrapolation with the hourglass design to enable generation beyond the training window size, thereby supporting long-sequence mesh synthesis.

Following MeshGPT~\cite{Meshgpt} and follow-up works~\cite{chen2024meshanythingv2, bpt}, mesh generation is formulated as a conditional sequence modeling task.
Given a 3D mesh, the process begins with a tokenizer that serializes the complex geometric and topological data into a discrete 1D sequence of tokens, denoted as $\mathcal{T} = (t_1, t_2, \dots, t_L)$, where $L$ indicates the sequence length.
This tokenization step bridges the gap between irregular 3D structures and standard sequence models.

To generate meshes, a Transformer-based decoder (GPT) learns to predict the sequence autoregressively.
Given a condition $\mathbf{c}$ (e.g., a point cloud), the model is trained to maximize the likelihood of the next token $t_i$ based on the preceding context $t_{<i}$.
The training objective is to minimize the standard cross-entropy loss over the dataset:
\begin{equation}
    \mathcal{L} = - \sum_{i=1}^{L} \log p(t_i \mid t_{<i}, \mathbf{c}; \theta)
    \label{eq:ar_loss}
\end{equation}
where $\theta$ represents the learnable parameters of the Transformer.
\section{Method}
\label{sec:method}
Given a set of 3D points as conditioning input, our goal is to generate an artist-style mesh with organized UV segmentation.
To achieve this, we propose Strips as Tokens (\name), a generative framework based on a unified strip-based representation. 
Our core contribution includes a serialization scheme that embeds macro-structural semantic cues like UV island boundaries into the token stream, and a stride-aware decoding protocol that allows the same model to generate both triangle and quadrilateral meshes.
The overview of the proposed framework is illustrated in Fig.~\ref{fig:ppl}.

We first describe our hierarchical geometry quantization process, which maps 3D coordinates into a compact discrete vocabulary (Sec.~\ref{sec:quant}). 
Next, we introduce our strip-based serialization, where meshes are converted into long, contiguous vertex streams with embedded UV transition markers (Sec.~\ref{sec:serialization}). 
Then, Sec.~\ref{sec:decoding} details our multi-topology interpretation protocol, which allows the recovered sequence to be adaptively decoded as either triangle or quad meshes. 
Finally, we discuss our three-stage training strategy, covering large-scale pretraining on triangles to fine-tuning on high-quality quad meshes (Sec.~\ref{sec:train}).

\subsection{Hierarchical Geometry Quantization}
\label{sec:quant}
We represent an artist mesh $\mathcal{M}$ as a tuple $(\mathcal{V}, \mathcal{F})$, where $\mathcal{V}$ is a set of $N$ vertices and $\mathcal{F}$ is a set of $M$ faces. 
Each vertex $v \in \mathcal{V}$ is defined by its 3D coordinates. 
In professional modeling workflows, these vertices are organized into polygons that follow specific structural rules, predominantly triangles and quadrilaterals. 
Accordingly, each face $f \in \mathcal{F}$ is defined as an ordered sequence of vertex indices, where the face degree $|f| \in \{3, 4\}$ denotes a triangle or a quadrilateral, respectively.

To bridge the gap between continuous geometric space and discrete tokens, we quantize the vertex coordinates onto a $512^3$ voxel grid following the three-level hierarchical strategy in DeepMesh~\cite{zhao2025deepmesh}.
Specifically, the mesh is normalized into a unit cube and each vertex is decomposed into a hierarchical tuple $(c_1, c_2, c_3)$ corresponding to $4^3$, $8^3$, and $16^3$ resolution levels (left of Fig.~\ref{fig:prefix}). 
Here $c_1 \in \mathcal{C}_{1}^{geo}$ identifies the coarsest grid cell, while $\{c_2, c_3\}$ specify the local relative position of the vertex within its respective parent cell from the previous level.
Together, this strategy provides the full $512^3$ precision, with $\mathcal{C}_{1}^{geo}$ serving as the coarsest coordinate codebook.
At this stage, the mesh geometry is fully discretized into a set of hierarchical tuples, but they remain unordered and detached from the topological faces $\mathcal{F}$.

\subsection{Strip-based Serialization}
\label{sec:serialization}
With the geometry discretized into hierarchical tokens, the remaining challenge is to establish a deterministic ordering that linearizes the mesh topology. 
Inspired by the concept of triangle strips~\cite{isenburg2001trianglestrip}, we propose to serialize the mesh into a sequence of vertices guided by the structural ``flow'' of adjacent faces. 
A strip is defined as a connected sequence of faces where each consecutive pair shares a common edge, a property that aligns perfectly with the organized edge-flow of artist meshes. 
By traversing the mesh through these shared-edge boundaries, we convert the graph-like connectivity of $\mathcal{F}$ into a coherent vertex stream $\mathcal{T}$.

\begin{algorithm}[t]
\caption{\textbf{Unified Strip Extraction (\name)}}
\label{alg:strip_extraction}
\begin{algorithmic}[1]

\Require Mesh faces $\mathcal{F}$, Stride parameter $\delta$ ($\delta=1$ for Triangle, $\delta=2$ for Quad).
\Ensure A set of extracted strips $\mathbf{S} = \{\mathcal{S}_1, \dots, \mathcal{S}_k\}$.

\Statex \textbf{Initialization.}
\State Build Edge-to-Face adjacency map $\textsc{E2F}$.
\State Initialize $\textsc{visited}[f]\gets \textbf{false}$ for all $f\in \mathcal{F}$.
\State Initialize strip list $\mathbf{S} \gets [\ ]$.

\Statex \textbf{Extraction Loop.}
\While{$\exists f \in \mathcal{F} \text{ s.t. } \textsc{visited}[f]=\textbf{false}$}
    \State \textbf{Start new strip.} $\mathcal{S}_{curr} \gets [\ ]$.
    \State Pick lowest unvisited face $f_{seed}$.
    \State $\mathbf{v} \gets \textsc{GetVertices}(f_{seed})$.
    
    % \Statex \hspace{\algorithmicindent}\textit{// Align winding order for Quad strips}
    \If{$\delta = 2$}
        \State Swap the last two vertices of $\mathbf{v}$.
    \EndIf
    
    \State Append $\mathbf{v}$ to $\mathcal{S}_{curr}$.
    \State Mark $\textsc{visited}[f_{seed}] \gets \textbf{true}$.
    \State Define boundary edge $e_{front} \gets (\mathbf{v}[-2], \mathbf{v}[-1])$.

    \Statex \hspace{\algorithmicindent}\textit{// Zipper-like growth}
    \While{\textbf{true}}
        \State $f_{next} \gets \textsc{NextFace}(\textsc{E2F}, e_{front}, \textsc{visited})$.
        \If{$f_{next} = \varnothing$}
            \State \textbf{break} \Comment{Hit boundary or visited face.}
        \EndIf
        
        \State $\mathbf{v}_{new} \gets \textsc{GetNewVertices}(f_{next}, e_{front})$.
        \Statex \Comment{Returns 1 $\mathbf{v}$ if $\delta=1$, pair of swapped $\mathbf{v}$ if $\delta=2$.}
        
        \State Append $\mathbf{v}_{new}$ to $\mathcal{S}_{curr}$.
        \State Mark $\textsc{visited}[f_{next}] \gets \textbf{true}$.
        \State Update $e_{front}$ based on $\mathbf{v}_{new}$.
    \EndWhile
    
    \State Append $\mathcal{S}_{curr}$ to $\mathbf{S}$.
\EndWhile
\State \Return $\mathbf{S}$.
\end{algorithmic}
% \vspace{-4mm}
\end{algorithm}
% \vspace{-4mm}

\subsubsection{Strip Extraction.}
\label{sec:method_strips}
We construct strips via a systematic ``zipper-like'' growth procedure that extracts topological paths from the input faces $\mathcal{F}$. 
As detailed in Alg.~\ref{alg:strip_extraction}, we first build an edge-to-face adjacency map and initialize all faces as unvisited. 
To extract a strip, we pick the first unvisited face (e.g., the faces were sorted by the lowest coordinate) as a seed and append its vertices to the output sequence.
\FF{The three vertices of the seed face are sorted by their coordinates, and the edge formed by the last two vertices in this order is designated as the initial boundary edge, which deterministically dictates the growth direction of the strip.}
\begin{wrapfigure}{r}{3.7cm}
\vspace{-1.5mm}
  \hspace*{-4mm}
  \centerline{
  \includegraphics[width=45mm]{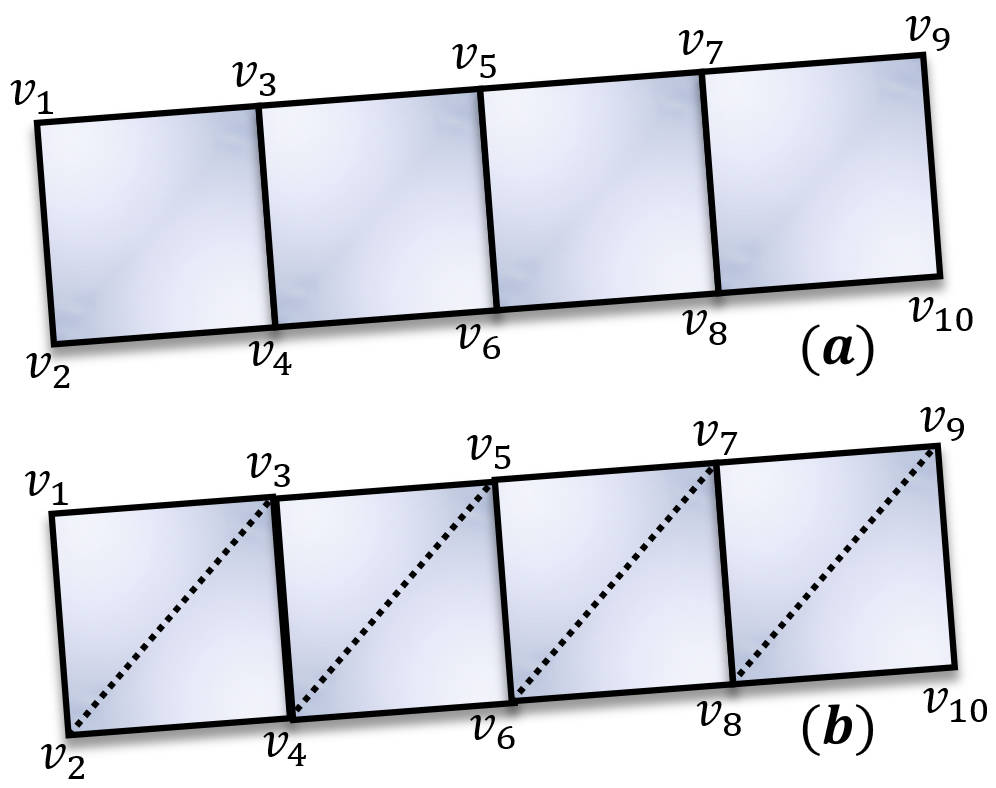}
  }
  \vspace*{-4mm}
\end{wrapfigure}
Starting from the seed face, the strip grows by repeatedly traversing across the current boundary edge to its adjacent unvisited face. 
This traversal is governed by a topology-specific stride $\delta$: in triangle mode ($\delta=1$), each step crosses a boundary edge to add a single new vertex, whereas in quad mode ($\delta=2$), each step crosses the edge to introduce a pair of new vertices.
This unified traversal ensures that both mesh types follow an identical ``grow-by-appending'' logic, where each step expands the sequence to induce a new face. 
To maintain this structural alignment, we enforce a consistent vertex ordering within each quadrilateral by swapping the last two indices of each face. 
As illustrated in the inset figure, this swap ensures that quad strips follow the same forward-moving order as triangle strips. 
The quad token sequence in inset figure~(a) is identical to the triangle token sequence in inset figure~(b).
By aligning the winding order in this manner, we achieve a structural consistency where both mesh types can be generated via a unified autoregressive flow.

The growth of a strip terminates when the current boundary edge either lies on the mesh boundary or connects only to faces that have already been visited. 
Once a strip reaches such a dead end, we select the next available unvisited face (following the same coordinate-based priority) as a new seed to initiate a subsequent strip. 
This process repeats iteratively until the entire face set $\mathcal{F}$ is covered, effectively decomposing the mesh into a collection of disjoint strips $\{\mathcal{S}_1, \mathcal{S}_2, \dots, \mathcal{S}_k\}$.
Crucially, this decomposition establishes a deterministic global vertex ordering.
\FF{We deliberately chose this greedy, lowest-coordinate-first strategy because it yields a fixed, spatially coherent traversal pattern that the network can learn easily. A globally optimized strip decomposition might reduce the total number of strips, but it could introduce erratic seed face locations and inconsistent traversal patterns, which hinders optimization in practice.}
By concatenating these strips and mapping each vertex to its corresponding hierarchical code, we could transform the complex mesh graph into a token sequence.

% To convert the decomposed mesh strips into a final token sequence, we concatenate the extracted strips $\{\mathcal{S}_j\}_{j=1}^k$ into a single sequence.
% Following the vertex ordering within each strip, every vertex $v_i$ is expanded into its hierarchical coordinate tuple $(c_{i,1}, c_{i,2}, c_{i,3})$.

\subsubsection{Strip Transition.}
% To delineate boundaries between strips, we follow DeepMesh~\cite{zhao2025deepmesh} by expanding the coarsest codebook $c_1$ with an additional special set of $4^3$ tokens $c_1^{sep}$ marked as \texttt{[SEP]}.
% Each token in this auxiliary set corresponds to the same spatial grid position as its original counterpart while additionally serving as a transition token to indicate the start of a new strip.
% When a new strip begins, its first vertex is encoded using these specialized $c_1^{sep}$ codes instead of the standard $c_1$. Therefore, it likely disrupts the existing prefix-sharing mechanism, resulting in longer token sequences.
% Fig.~\ref{fig:prefix} shows a clearer toy example to help understand how prefix sharing works and why transitions may will break prefix sharing.
To serialize these disjoint strips into a unified token stream, we require a mechanism to explicitly delineate topological boundaries.
Following the codebook expansion strategy introduced in DeepMesh~\cite{zhao2025deepmesh}, we distinguish strip boundaries by expanding the vocabulary of the coarsest codebook level $\mathcal{C}_1^{geo}$.
Specifically, we augment the coarsest codebook level $\mathcal{C}_1^{geo}$ with a separate parallel set of tokens, denoted as $\mathcal{C}_1^{t}$.
Each token in this auxiliary set corresponds to the same spatial grid position as its standard counterpart but serves a distinct semantic role.
During sequence construction, the first vertex of every new strip is encoded using these specialized $\mathcal{C}_1^{t}$ tokens.
In this way, we effectively embed the ``start-of-strip'' signal directly into the geometric sequence.
This avoids the need for inserting separate delimiter tokens, ensuring that the explicit boundary definition does not increase the overall sequence length.

\begin{figure}[!tp]
    \centering
    \begin{overpic}[width=\linewidth]{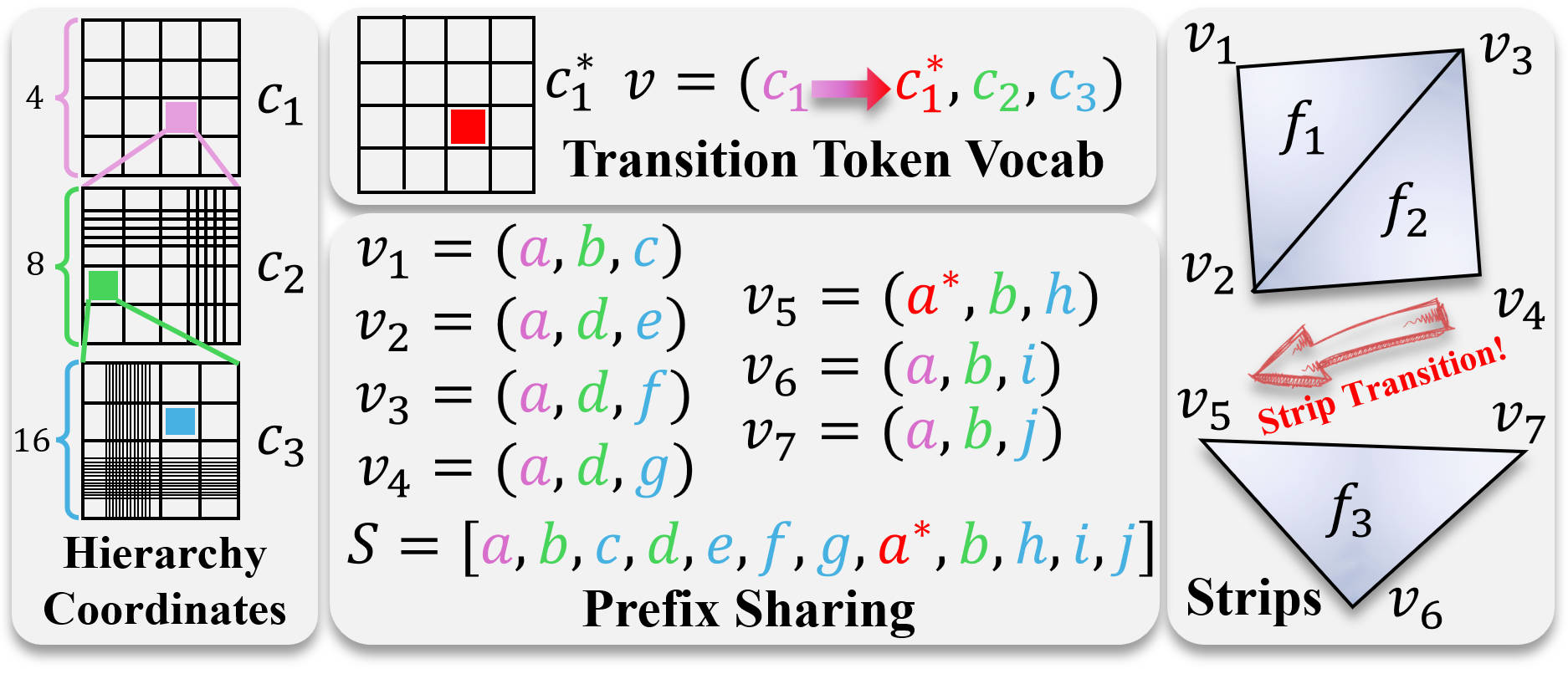}
    \end{overpic}
    \vspace{-8mm}
    \caption{\textbf{\FF{Mesh tokenization with prefix sharing.}}  We use a three-level hierarchical coordinate $(c_1,c_2,c_3)$ with prefix sharing to compress token sequences following DeepMesh~\cite{zhao2025deepmesh}, repeated prefixes between consecutive vertices are omitted.
    To mark a strip transition, we introduce a special top-level token vocabulary $c_1^{*}$ (red), which is distinct from $c_1$ but serves the same role. Whenever $c_1^{*}$ appears, the prefix sharing state is reset, starting a new prefix context.
}
    \label{fig:prefix}
    \vspace{-4mm}
\end{figure}

\begin{figure}[!tp]
    \centering
    \begin{overpic}[width=\linewidth]{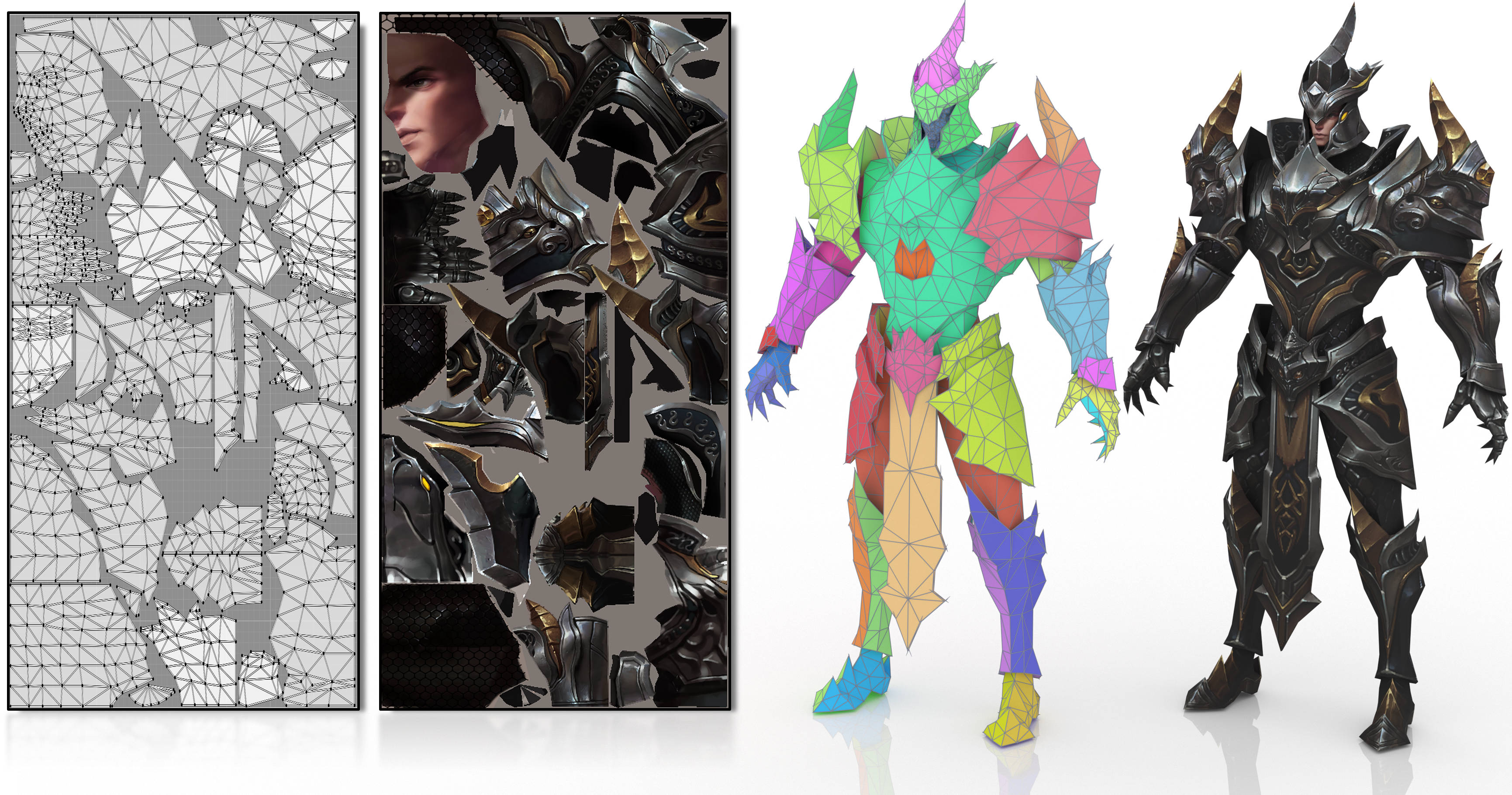}
    \put(5,  0.7){\textbf{UV Map}}
    \put(30, 0.7){\textbf{Texture}}
    \put(50, 0.7){\textbf{UV Segment}}
    \put(77, 0.7){\textbf{Artist Mesh}}
    \end{overpic}
    \vspace{-8mm}
    \caption{\textbf{Artist-created meshes with UV chart partitions.} We split artist meshes into UV parts and let \name traverse all triangles within one part before a UV  segmentation transition to the next part, enabling native UV segmentation during generation.
}
    \label{fig:uvseg}
    \vspace{-4mm}
\end{figure}
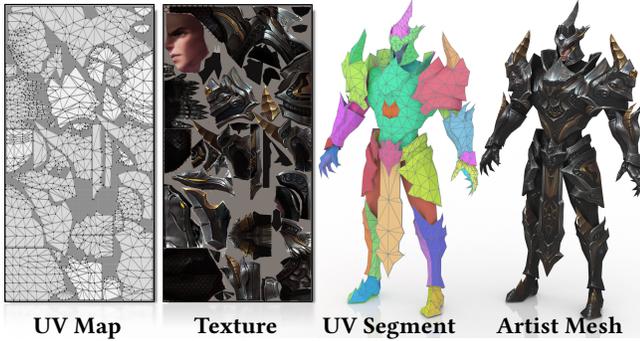
\subsubsection{UV Segmentation.}
\label{sec:method_uvseg}
Beyond strip connectivity, our tokenizer natively supports UV segmentation to preserve the macro-structural organization of artist meshes.
As illustrated in Fig.~\ref{fig:uvseg}, we partition the mesh faces into disjoint groups based on their UV islands and impose a deterministic traversal order across these islands (bottom to up).
Within each island, the strip-based encoding proceeds as usual, with the constraint that the next seed face must be selected from the current island until all its constituent faces are exhausted.
To distinguish these semantic boundaries, we further expand the coarsest codebook $\mathcal{C}_1^{geo}$ with an additional set $\mathcal{C}_1^{uv}$, which denote the completion of a UV island and a transition to the next UV segmentation.
Notably, the $\mathcal{C}_1^{uv}$ strictly subsumes the function of $\mathcal{C}_1^{t}$: it signals both the termination of a strip and a higher-level switch between distinct UV charts.
By injecting these artist-preferred semantic cues into the sequence, we enable the model to learn not only the surface geometry but also the high-level layout intent inherent in professional mesh modeling.
\FF{Note that our model learns only the UV chart partitioning (i.e., which faces belong to which island), not the UV coordinates themselves; a standard unwrapping algorithm in Blender~\cite{Blender} is applied afterward to compute the actual 2D parameterization from the predicted segmentation.} 

% Consequently, the final sequence $\mathcal{T}$ maintains a compact form where each vertex $v_i$ is represented by its hierarchical tokens $(c_{i,1}, c_{i,2}, c_{i,3})$.
Consequently, the final sequence $\mathcal{T}$ remains compact, with each vertex $v_i$ encoded by its hierarchical tokens $(c_{i,1}, c_{i,2}, c_{i,3})$.
While higher-level tokens remain standard, the first-level token $c_{i,1}$ is drawn from an augmented vocabulary $\mathcal{C}_{i}^{*}$ that integrates spatial, structural, and semantic information: 
\begin{equation}
    c_{i,1} \in \mathcal{C}_1^* = \underbrace{\mathcal{C}_{1}^{geo}}_{\text{Standard}} \cup \underbrace{\mathcal{C}_1^{t}}_{\text{Strip Transition}} \cup \underbrace{\mathcal{C}_1^{uv}}_{\text{UV Segmentation}}
    \label{eq:expanded_c1}
\end{equation}
Under this scheme, a typical vertex stream $\mathcal{T}$ appears as:
\begin{equation}
    \mathcal{T} = \big( \dots, \underbrace{(c_{i,1}, c_{i,2}, c_{i,3})}_{\text{Standard}}, \dots, \underbrace{(c_{j,1}^{t}, c_{j,2}, c_{j,3})}_{\text{New Strip}}, \dots, \underbrace{(c_{k,1}^{uv}, c_{k,2}, c_{k,3})}_{\text{New UV Island}} \big).
    \label{eq:actual_sequence}
\end{equation}
This unified format ensures that the serialization is natively aware of the mesh's macro-structural organization.
% where each structural transition $c^*$ automatically triggers a reset of the prefix sharing context and the topological frontier. 
% This implicit signaling ensures that the serialization is both spatially efficient and natively aware of the mesh's macro-structural organization.

Finally, inspired by DeepMesh~\cite{zhao2025deepmesh}, we employ a prefix sharing strategy to minimize sequence length by exploiting the inherent spatial continuity within each strip.
Consecutive vertices often share identical coarse locations $c_1$ or parent cells $c_2$.
In such cases, we omit the redundant prefixes.
For instance, if a vertex $v_{i+1}$ and its preceding vertex $v_i$ share the same $c_1$ and $c_2$ codes, the original sequence $[(c_{i,1}, c_{i,2}, c_{i,3}), (c_{i+1,1}, c_{i+1,2}, c_{i+1,3})]$ is compressed into $[c_{i,1}, c_{i,2}, c_{i,3}, c_{i+1,3}]$.
In this case, the representation of $v_{i+1}$ is reduced from a three-token tuple to a single token.
Crucially, structural tokens from $\mathcal{C}_1^{t}$ and $\mathcal{C}_1^{uv}$ serve as absolute synchronization points.
They are never compressed and implicitly force a reset of the sharing context, ensuring that topological transitions remain explicit and unambiguous to the model.
Fig.~\ref{fig:prefix} shows a clear toy example to help understand how the token sequence is obtained.
\FF{Empirically, on our test set the token distribution across levels is $c_1$: 20.7\%, $c_2$: 35.0\%, $c_3$: 44.3\%, confirming that prefix sharing effectively compresses the majority of vertices to one or two tokens.}

\begin{figure}[!tp]
    \centering
    \begin{overpic}[width=\linewidth]{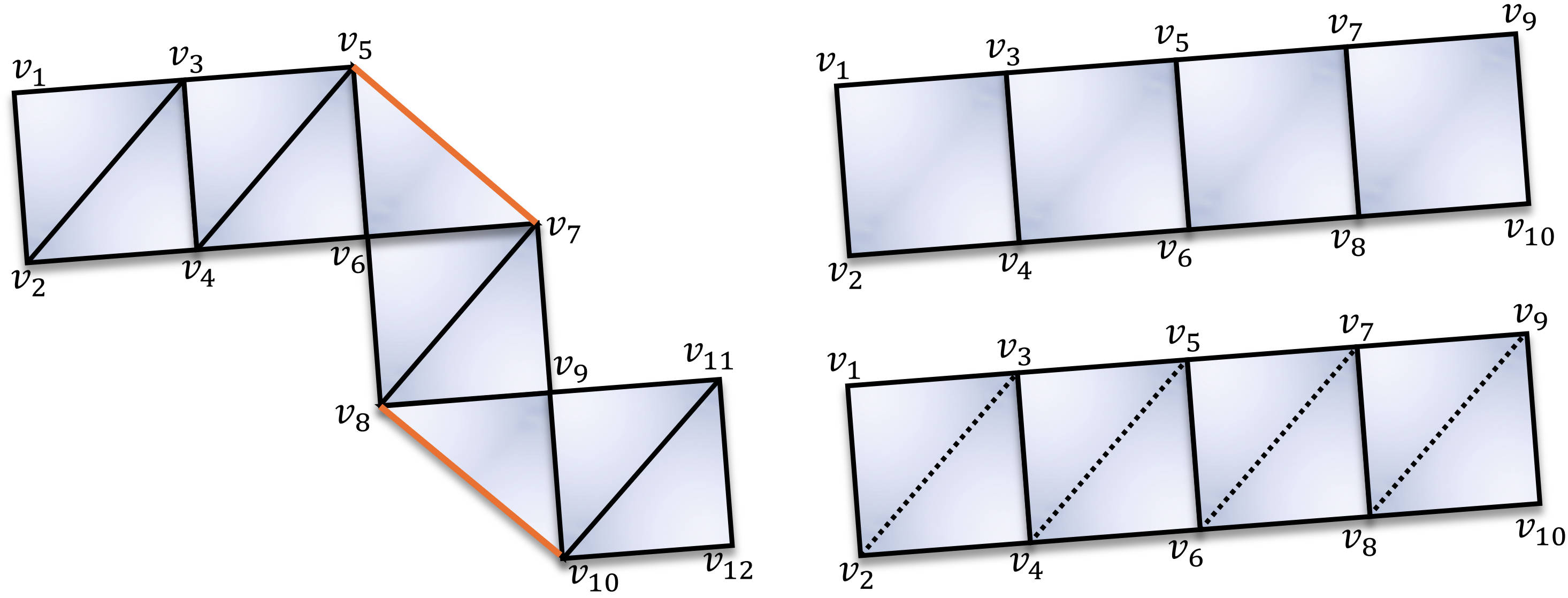}
    \put(16, 15){\textbf{(a)}}
    \put(92, 20){\textbf{(b)}}
    \put(92, 0){\textbf{(c)}}
    \end{overpic}
    \vspace{-7mm}
    \caption{\textbf{Unified representation of triangle and quad using strips.}
    Triangle strips may locally ``turn'' under edge flips (a).
In contrast, quad strips avoid this ambiguity (b), as each step admits only a single forward direction.
Moreover, sequences tokenized from a quad mesh can be decoded into triangles while still preserving high quality (c). Note that the quad token sequence of (b) is totally the same as the triangle token sequence of (c).
}
    \label{fig:quadstrip}
    \vspace{-4mm}
\end{figure}
\subsubsection{Properties of the Representation.}
The proposed strip-based serialization offers three fundamental advantages for mesh generative modeling. 

First, by capturing the long-range structural ``flow'' typical of artist meshes, our representation provides a stronger inductive bias for learning regular topology and consistent connectivity compared to randomized or patch-based orderings. 

Second, our unified stride-based formulation enables topological synergy between disparate mesh types; by linearizing triangles and quadrilaterals into the same vertex stream, we allow the model to share geometric priors across different domains.
Furthermore, training on quad meshes can improve the quality of triangle-strip sequences.
As shown in Fig.~\ref{fig:quadstrip}~(a), triangle strips on certain artist meshes may exhibit occasional ``turns.'' 
While such turns will not affect the quality of the generation model, they introduce additional ordering variability, forcing the model to learn stronger traversal priors.
Quad strips largely avoid this issue (Fig.~\ref{fig:quadstrip}~(b)): large-angle turns are rare within the quadrilateral zone, so the traversal naturally progresses forward and rarely produces large-angle bends.

\begin{figure}[!tp]
    \centering
    \begin{overpic}[width=\linewidth]{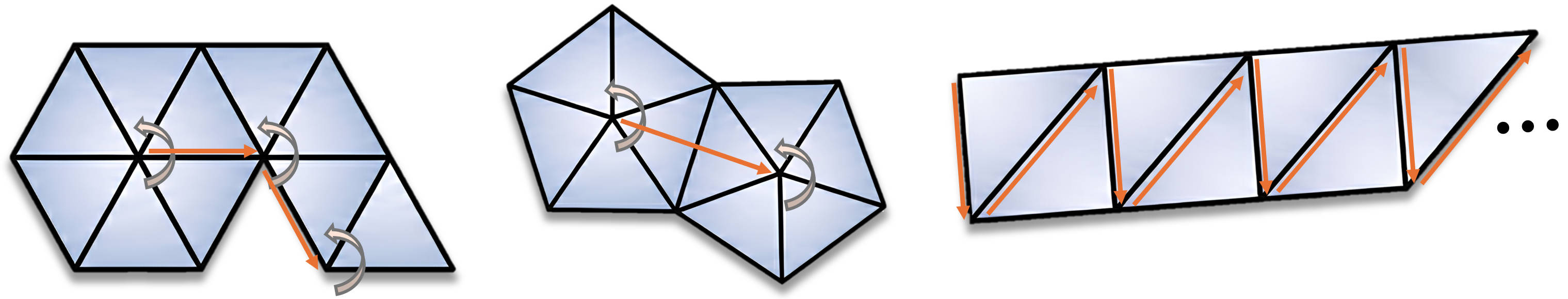}
    \put(10, -2.5){\textbf{BPT}}
    \put(37, -2.5){\textbf{DeepMesh}}
    \put(73, -2.5){\textbf{Ours}}
    \end{overpic}
    \vspace{-5mm}
    \caption{\textbf{Different face ordering defined by other methods.}
BPT~\cite{bpt} and DeepMesh~\cite{zhao2025deepmesh} traverse local fan-/disk-shaped neighborhoods, i.e., triangles rotate around a vertex, which triggers patch transitions more frequently.
In contrast, our strip-based ordering can, in principle, extend arbitrarily long.
}
    \label{fig:tkizer}
    \vspace{-4mm}
\end{figure}

Third, our approach significantly optimizes encoding efficiency relative to patch-based methods like DeepMesh~\cite{zhao2025deepmesh}. 
As illustrated in Fig.~\ref{fig:tkizer}, patch-based methods partition the mesh into numerous small fragments (typically 5--7 faces each), each necessitating a transition token that resets the prefix sharing context. 
In contrast, our decomposition into long, contiguous strips drastically reduces the frequency of these resets, allowing spatial continuity to persist over larger spans and effectively amortizing the transition overhead to produce a more concise serialization.
We report the average compression ratio achieved by different tokenizers on 100 randomly sampled meshes from Objaverse~\cite{objaverseXL} in Table~\ref{tab:compRate}.
\FF{Despite using a slightly larger vocabulary than DeepMesh~\cite{zhao2025deepmesh} (the additional tokens are used to support UV segmentation), our tokenizer attains a noticeably higher compression ratio, indicating a more efficient sequence representation under the same discrete budget.}

\begin{table}[!tp]
\centering
\caption{\textbf{Comparison of vocabulary size and average compression rate.} The compression rate is computed as the token sequence length divided by (face count × 9).}
\vspace{-3mm}
\resizebox{0.7\columnwidth}{!}{
    \begin{tabular}{c|ccc}
    \toprule
Metrics              & BPT & DeepMesh & SATO  \\ \midrule

Vocab Size  $\downarrow$           & 40960       & 4736           & 4800 \\
Comp Rate  $\downarrow$            & 0.228      & 0.330          & 0.283 \\
\bottomrule
\end{tabular}
}
\label{tab:compRate}
\vspace{-3mm}
\end{table}

\begin{figure*}[!tp]
    \centering
    \begin{overpic}[width=\linewidth]{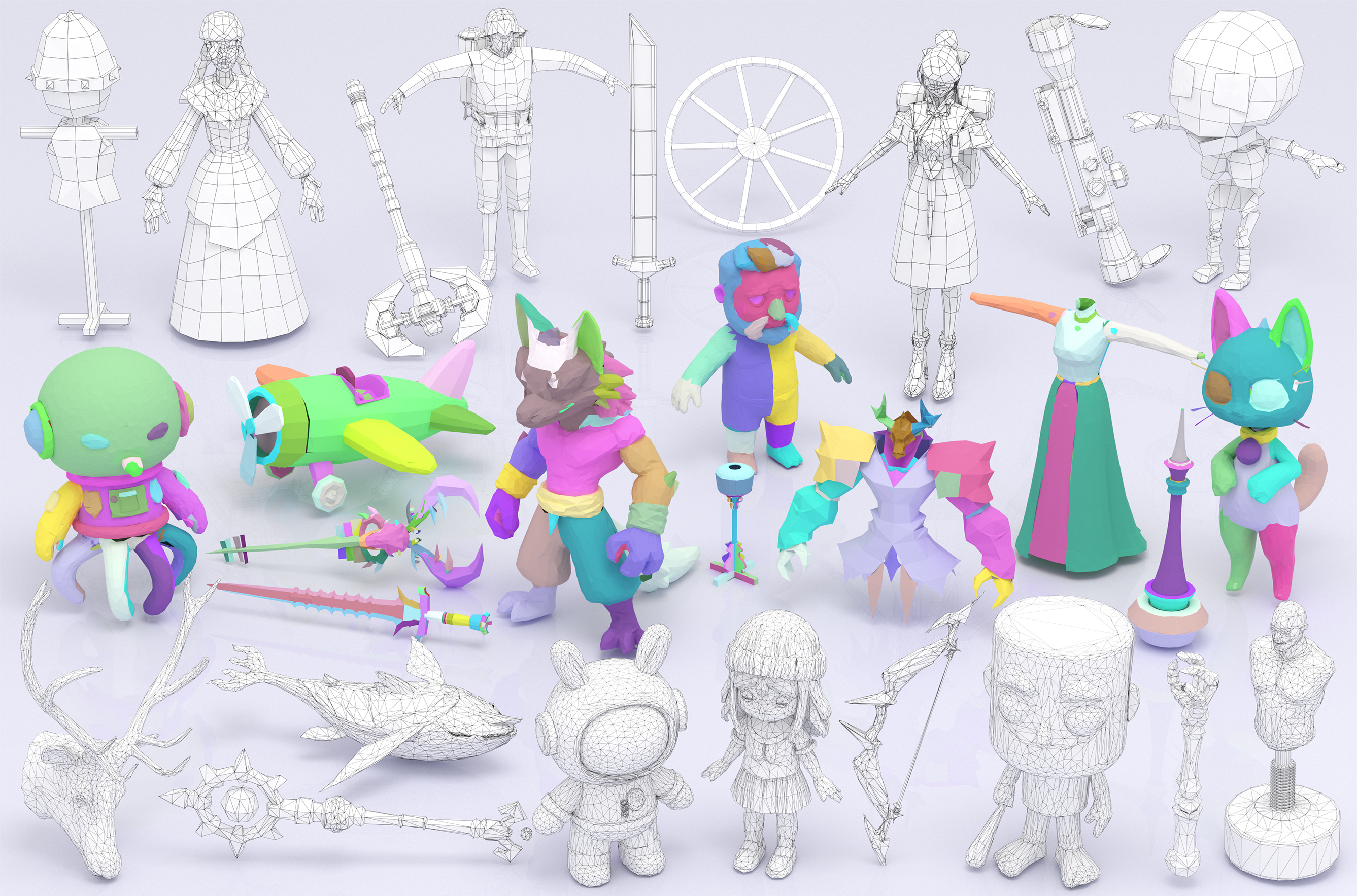}
    \end{overpic}
    \vspace{-8mm}
    \caption{\textbf{The gallery of \name illustrates our model’s outputs across three tasks.} From bottom to top, it shows triangular mesh generation, shape generation with UV segmentation, and quadrilateral mesh generation. SATO supports all three tasks within a single framework and achieves compelling results on each of them.
}
    \label{fig:gallery}
    \vspace{-4mm}
\end{figure*}

\subsection{Topology-Specific Decoding}
\label{sec:decoding}
The conversion from the token sequence $\mathcal{T}$ back to a mesh $\mathcal{M}$ is governed by a deterministic decoding protocol. 
For each vertex $v_i$, the decoder first restores the full hierarchical coordinates: if the input is a compressed residual $c_{i,3}$, it is prepended with the $(c_{i,1}, c_{i,2})$ prefix cached from the preceding vertex following DeepMesh~\cite{zhao2025deepmesh}. 
The global mesh structure is managed by structural markers embedded within the $\mathcal{C}_1^*$ vocabulary. 
Specifically, $\mathcal{C}_1^{t}$ signals the termination of the current strip and the initiation of a new one, while $\mathcal{C}_1^{uv}$ indicates a transition between disjoint UV islands. 
Upon detecting either marker, the decoder immediately resets both the coordinate cache and the topological frontier, ensuring that the subsequent vertices are interpreted as a fresh seed face for the next segment.

The primary distinction of our protocol is its support for multi-topology recovery via an adjustable vertex stride $\delta \in \{1, 2\}$. 
Leveraging the consistent vertex ordering enforced during the encoding stage, the decoder can interpret a single geometric stream through disparate topological rules. 
In triangle-mesh mode ($\delta=1$), each successive vertex $v_{i+2}$ completes a face $f_i = (v_i, v_{i+1}, v_{i+2})$. 
In quadrilateral-mesh mode ($\delta=2$), the decoder processes vertices in pairs; for any two newly generated vertices $(v_{2i+2}, v_{2i+3})$, it assembles a quad face $q_i = (v_{2i}, v_{2i+1}, v_{2i+3}, v_{2i+2})$. 
For example, a six-vertex sequence $(v_0, \dots, v_5)$ is interpreted as four triangles under $\delta=1$, or as two quadrilaterals $q_0 = (v_0, v_1, v_3, v_2)$ and $q_1 = (v_2, v_3, v_5, v_4)$ under $\delta=2$. 
This unified interpretive framework enables the same autoregressive model to learn shared geometric priors across heterogeneous datasets by simply toggling the decoding stride.
\FF{Notably, switching between triangle and quad output requires no special tokens or architectural changes; the user simply sets $\delta$ at inference time.
In quad mode, if a strip contains an odd number of vertices after the seed face, the final unpaired vertex is decoded as a triangle.
The detokenizer also strictly discards structurally invalid markers (e.g., consecutive $\mathcal{C}_1^{t}$ or $\mathcal{C}_2$ tokens without intervening geometry tokens); in practice, this failure mode has never been observed with our trained model.
Additionally, vertices from different strips that share the same quantized coordinates within a UV region are welded during decoding to ensure a connected mesh.}

\subsection{Training with \name}
\label{sec:train}

The training pipeline of \name is organized into three stages: (i) large-scale triangle-mesh pretraining, (ii) UV-segmentation post-training, and (iii) quad-mesh fine-tuning.
% This staged design is primarily motivated by optimization difficulty and data availability.

\subsubsection{Model Architecture and Optimization}
\name uses a $0.5$B parameter autoregressive hourglass transformer backbone, which has been shown to be well-suited for mesh generation~\cite{hao2024meshtron}.
\FF{Specifically, the transformer consists of 21 layers with 8 attention heads and 1024-dimensional embeddings.}
For point cloud conditioning, instead of using a pretrained and frozen point cloud VAE encoder as in prior work~\cite{zhao2025deepmesh}, we adopt the same VAE architecture as Hunyuan3D~\cite{lei2025hunyuan3dstudio} but train it from scratch after reducing the layers and token length to better align with inputs; the resulting encoder has roughly $0.27$B parameters.
\FF{Concretely, we reduce the decoder from 16 layers to 12 layers and the condition token count from 4096 to 1024 to better match our point cloud inputs.}
We optimize the model using the standard cross-entropy loss. Due to the high resolution of our tokenization, mesh sequences often exceed the attention window of the Transformer. To address this, we adopt the truncated-window training strategy~\cite{zhao2025deepmesh, hao2024meshtron} with 9K window size, where the model is trained on overlapping segments of the full sequence. \FF{Specifically, during each training iteration, we randomly select a contiguous subsequence of 9K tokens from the full mesh token stream as the training input.} This allows \name to maintain local geometric coherence while scaling to complex meshes with large token counts.

\subsubsection{Data Processing}
\label{sec:dataprocess}
Removing noisy or low-quality samples from millions of training shapes is essential to mesh generation training.
We apply the following filtering pipeline to construct our dataset.
\FF{For all meshes, we first discard non-manifold models and merge duplicate vertices.
We then keep shapes whose face count lies in $[500, 16000]$ and whose vertex-to-face ratio does not exceed $1.0$; models violating the latter criterion are often highly fragmented and close to a triangle soup.}
All data are randomly rotated along the Z-axis at four angles $[0, 90, 180, 270]$ \FF{before tokenization}.
% Finally, we render each model and use GPT\FF{need to ensure} to assign a semantic label; samples without a clear, recognizable semantic category are excluded.
For UV-related training, we additionally validate UV segmentation and keep only models whose number of UV islands lies in $[10, 300]$ to avoid excessively fragmented UV layouts.

\subsubsection{Training.}
Then we train our \name network with three stages.
\paragraph{Stage I: Triangle Mesh Pretraining.}
We first train the backbone Transformer together with the base \name tokenizer without UV segmentation on a large corpus of triangle mesh datasets.
This stage establishes strong geometric priors, including local strip continuation patterns and the alignment between mesh tokens and the conditioning point clouds.
Empirically, such priors are crucial for stable autoregressive training.

\paragraph{Stage II: UV Segmentation Post-Training.}
Directly training a UV segmentation model from scratch is challenging.
In early training, the model must simultaneously (a) learn the basic correspondence between mesh sequences and the conditioning input, and (b) discover higher-level semantic structure induced by UV islands (including predicting segment boundaries and handling inter-island transitions).
These objectives interact and often lead to slow convergence or degenerate solutions during our test (Sec.~\ref{sec:abla_uv_pretrain}).
To mitigate this, we perform a second-stage post-training where we initialize from the pretrained triangle model and then introduce the UV segmentation tokenization described in Sec.~\ref{sec:method_uvseg}.
In this stage, the model mainly adapts to the newly injected segmentation tokens (e.g., $\mathcal{C}_1^{uv}$) and the corresponding inter-island transition rules, while retaining the learned geometric and conditioning alignment from Stage~I. This strategy can significantly accelerate the convergence of the UV segmentation module and improve its performance.

\paragraph{Stage III: Quad Mesh Fine-tuning.}
High-quality quad meshes are substantially less abundant than triangle meshes, making it impractical to train an autoregressive quad generator from scratch at scale.
Thanks to the compatibility of our strip-based representation, we fine-tune the model initialized from Stage~I/II using the quad-strip decoding rule in Sec.~\ref{sec:method_strips}.
This transfers the majority of the learned priors from the triangle domain and only requires a relatively small quad-mesh dataset to adapt the model to quad-specific connectivity and strip statistics, while also allowing quad fine-tuning to modestly feed back and improve triangle generation quality (Sec.~\ref{sec:quadabla}).

\begin{table*}[!t]
% \vspace{-2mm}
\caption{\textbf{Quantitative comparison on ShapeNet~\citep{chang2015shapenet}, Thingi10K~\citep{zhou2016thingi10k}, and Objaverse~\citep{objaverseXL} datasets.}
The \underline{\textbf{best}} scores are emphasized in bold with underlining, while the \textbf{second best} scores are highlighted only in bold.
}
\vspace{-3mm}
\label{tab:comp_all}
\centering
\resizebox{.9\linewidth}{!}{
\begin{tabular}{l|cccc|cccc|cccc}
\toprule
& \multicolumn{4}{c|}{\textbf{ShapeNet}} & \multicolumn{4}{c|}{\textbf{Thingi10K}} & \multicolumn{4}{c}{\textbf{Objaverse}} \\ \midrule
\textbf{Method}
& $\mathrm{NC}\uparrow$ & $\mathrm{CD}\downarrow$ & $\mathrm{HD}\downarrow$ & $\mathrm{F1}\uparrow$
& $\mathrm{NC}\uparrow$ & $\mathrm{CD}\downarrow$ & $\mathrm{HD}\downarrow$ & $\mathrm{F1}\uparrow$
& $\mathrm{NC}\uparrow$ & $\mathrm{CD}\downarrow$ & $\mathrm{HD}\downarrow$ & $\mathrm{F1}\uparrow$ \\
\midrule
MeshAnythingV2~\cite{chen2024meshanythingv2} & 0.911 & 0.009 & 0.078 & 0.361 & 0.841 & \textbf{0.022} & 0.168 & 0.162 & 0.858 & \textbf{0.016} & \underline{\textbf{0.117}} & 0.208 \\
TreeMeshGPT~\cite{lionar2025treemeshgpt}    & 0.840 & 0.034 & 0.161 & 0.439 & 0.791 & 0.058 & 0.228 & 0.236 & 0.783 & 0.057 & 0.238 & 0.188 \\
BPT~\cite{bpt}            & 0.962 & \textbf{0.003} & \underline{\textbf{0.017}} & \textbf{0.605} & \textbf{0.874} & 0.028 & \underline{\textbf{0.141}} & \textbf{0.248} & 0.841 & 0.030 & 0.137 & \textbf{0.265} \\
DeepMesh~\cite{zhao2025deepmesh}       & \textbf{0.967} & 0.004 & 0.037 & 0.532 & 0.853 & 0.026 & 0.167 & 0.157 & \textbf{0.859} & 0.020 & \textbf{0.120} & 0.240 \\
\textbf{\name}          & \underline{\textbf{0.975}} & \underline{\textbf{0.002}} & \textbf{0.032} & \underline{\textbf{0.807}}
              & \underline{\textbf{0.916}} & \underline{\textbf{0.009}} & \textbf{0.154} & \underline{\textbf{0.460}}
              & \underline{\textbf{0.909}} & \underline{\textbf{0.009}} & \underline{\textbf{0.117}} & \underline{\textbf{0.503}} \\
\bottomrule
\end{tabular}
}
\vspace{-3mm}
\end{table*}

\section{Experimental Results}
\name supports three tasks within a single framework: triangular mesh generation, UV segmentation generation, and quadrilateral mesh generation. Fig.~\ref{fig:gallery} presents a gallery of our representative outputs (bottom to top): generated triangular meshes, generated UV segmentation (shown with color encoding), and generated quadrilateral meshes, highlighting the strong generative capability of our model.

\begin{figure}[!tp]
    \centering
    \begin{overpic}[width=\linewidth]{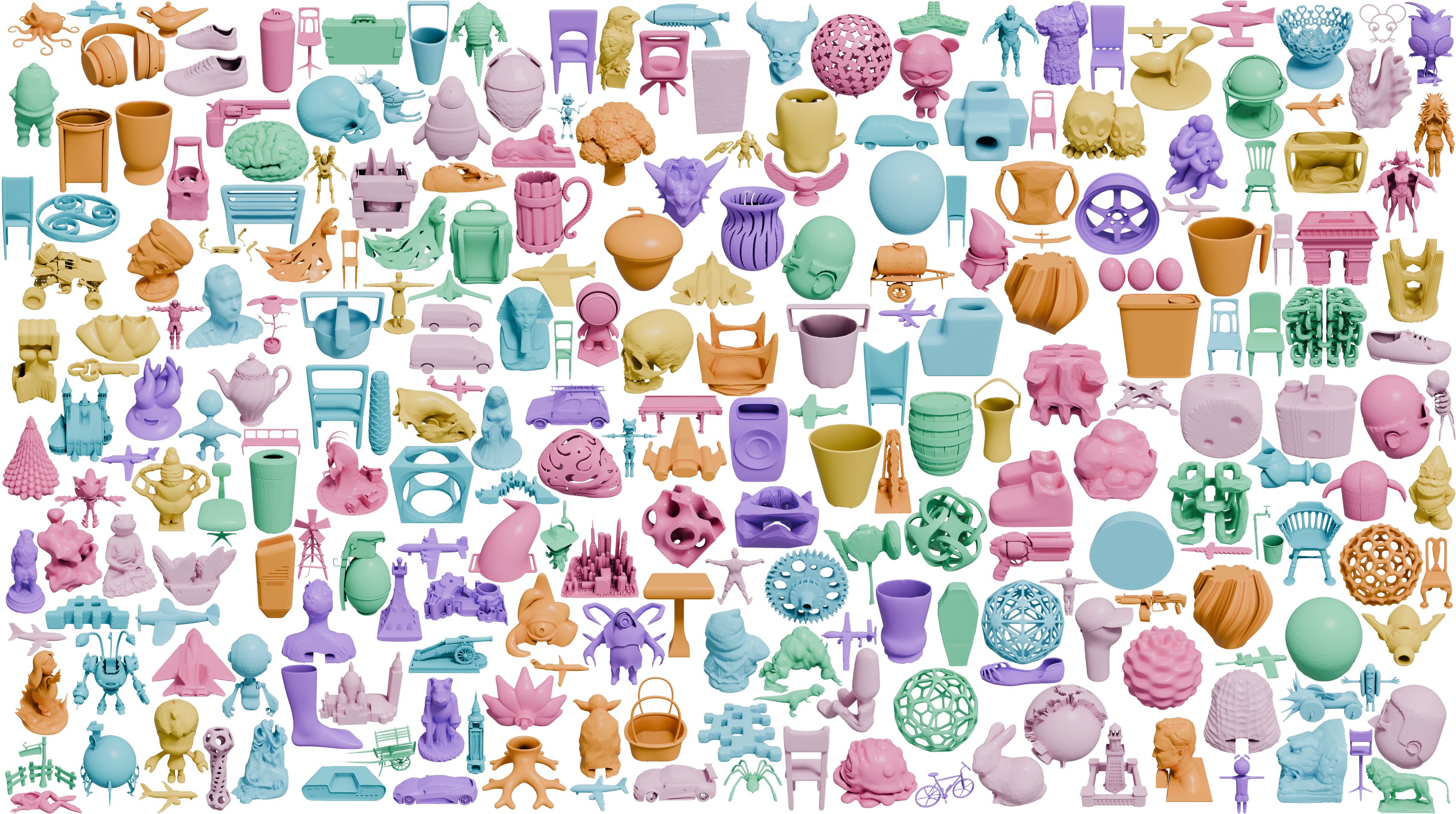}
    \end{overpic}
    \vspace{-8mm}
    \caption{\FFF{\textbf{Overview of our test dataset.} 250 models selected from the ShapeNet~\cite{chang2015shapenet}, Thingi10K~\cite{zhou2016thingi10k} and Objaverse~\cite{objaverseXL} dataset.}}
    \label{fig:testdataset}
    \vspace{-4mm}
\end{figure}

\begin{figure*}[!tp]
    \centering
    \begin{overpic}[width=\linewidth]{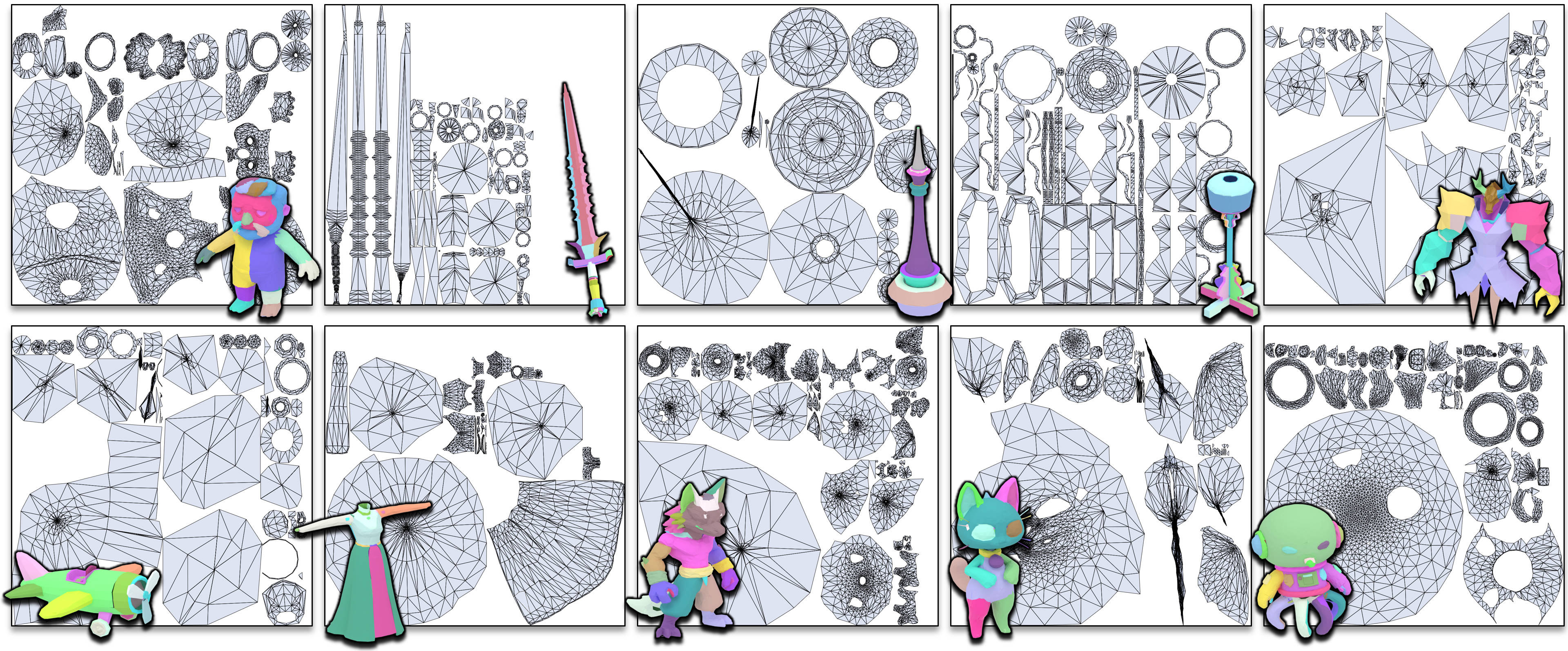}
    \end{overpic}
    \vspace{-9mm}
    \caption{\textbf{Gallery of UV unwrapping results using our generated UV segmentation.} The shapes are taken from Fig.~\ref{fig:gallery}, and both UV unwrapping and visualization are obtained through the unwrapping algorithm in Blender~\cite{Blender}.
}
    \label{fig:UVgallery}
    \vspace{-2mm}
\end{figure*}

\paragraph{Implementation Details}
Our curated artist mesh dataset is aggregated from Objaverse~\cite{objaverseXL}, ShapeNet~\cite{chang2015shapenet}, Thingi10K~\cite{zhou2016thingi10k} and licensed datasets \FFF{from 
Shutterstock~\cite{shutterstock}. 
The model in Fig.~\ref{fig:teaser} using asset by 
SDragonXF on Sketchfab~\cite{dragonhead3}.
And the shapes in Fig.~\ref{fig:artist},~\ref{fig:ppl},~\ref{fig:uvseg},~\ref{fig:gallery},~\ref{fig:UVgallery},~\ref{fig:app_uv} and ~\ref{fig:diversity} are from Shutterstock~\cite{shutterstock}.
}
After the preprocessing in Sec.~\ref{sec:dataprocess}, we obtain about $1.47M$ triangle meshes, among which $1.11M$ include high-quality UV chart partitions.
We additionally collect $120K$ UV-annotated quad meshes for fine-tuning. 
For each mesh, we randomly sample $81920$ points as the point cloud condition.
We train our model in three stages.
First, we pre-train the triangle-mesh generator on $64$ NVIDIA A800 GPUs for \FF{approximately 200K steps ($\sim$7 days)}.
Then post-train the model on UV-segmented data using $256$ A800 GPUs for \FF{approximately 80K steps ($\sim$3 days)} to enable native UV-aware generation.
Finally, we fine-tune the model on a high-quality quad dataset using $64$ A800 GPUs for \FF{approximately 25K steps ($\sim$1 day)}.
For both pre-training and post-training, we use a cosine learning-rate schedule decaying from $10^{-4}$ to $10^{-5}$; for quad mesh fine-tuning, we fix the learning rate to $10^{-5}$.
During training, we randomly sample a contiguous subsequence of length 9K tokens from each full token stream as the training input.
At inference time, we enable KV-cache throughout autoregressive decoding and use temperature sampling with $T=0.5$.

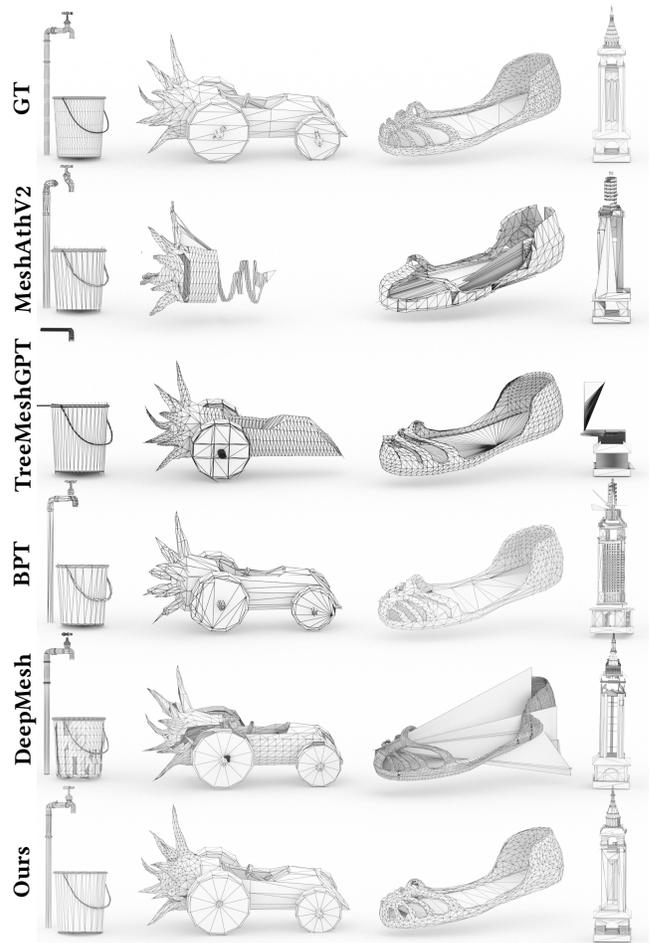
\begin{figure}[!tp]
    \centering
    \begin{overpic}[width=0.95\linewidth]{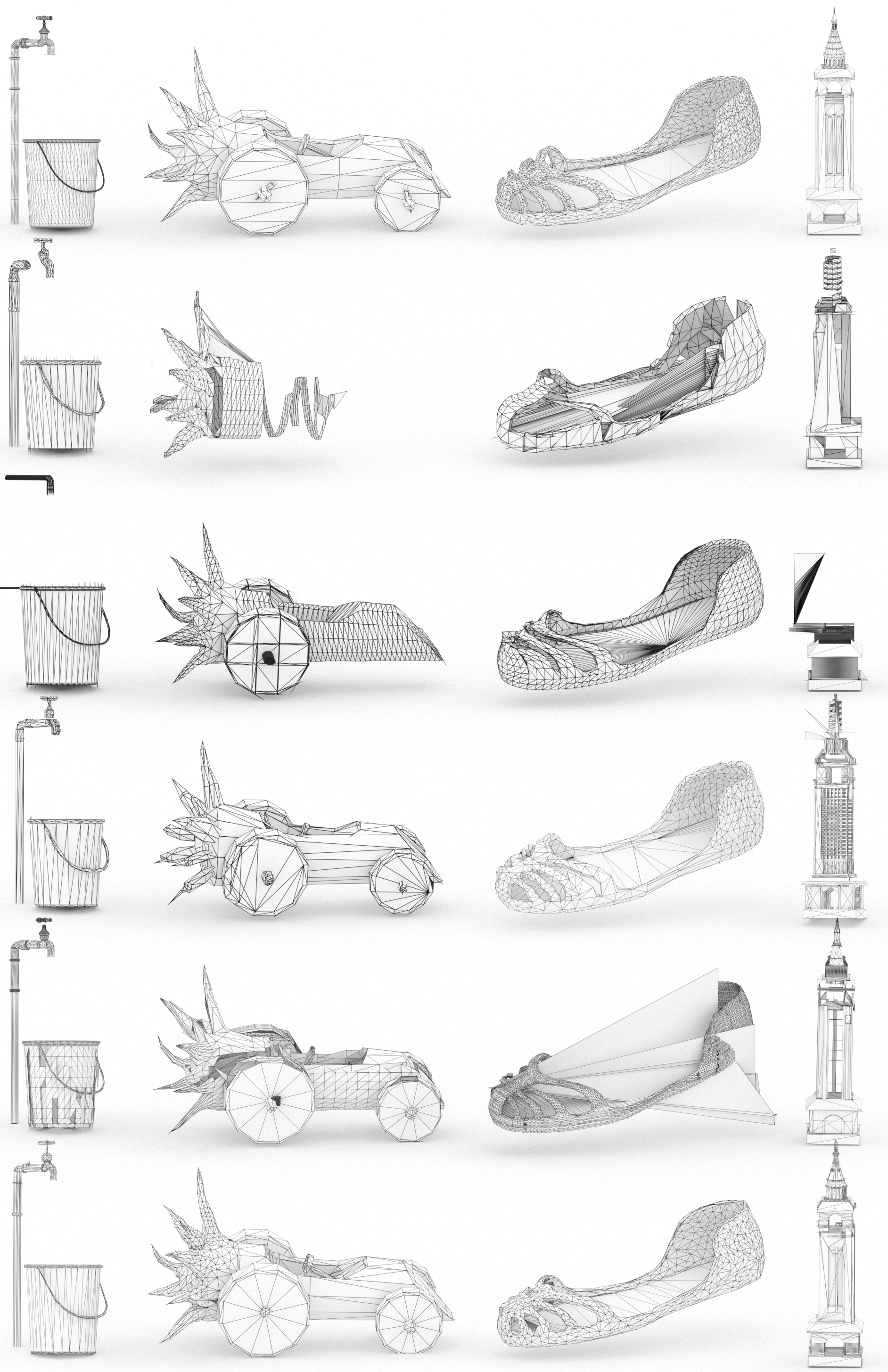}
    \put(-2.5, 88){\rotatebox{90}{\textbf{GT}}}
    \put(-2.5, 67){\rotatebox{90}{\textbf{MeshAthV2}}}
    \put(-2.5, 48){\rotatebox{90}{\textbf{TreeMeshGPT}}}
    \put(-2.5, 38){\rotatebox{90}{\textbf{BPT}}}
    \put(-2.5, 19){\rotatebox{90}{\textbf{DeepMesh}}}
    \put(-2.5, 5 ){\rotatebox{90}{\textbf{Ours}}}
    \end{overpic}
    \vspace{-4mm}
    \caption{\textbf{Qualitative comparison with baseline methods across different shapes.} Our approach consistently produces high-quality artist meshes with stable structure and clean surface.}
    \label{fig:comp1}
    \vspace{-4mm}
\end{figure}

\subsection{Triangle Mesh Generation}
\label{sec:Tricomparison}

\paragraph{Approaches}
We include 4 state-of-the-art (SOTA) methods for comparison: MeshAnythingV2~\citep{chen2024meshanythingv2}, BPT~\citep{bpt}, TreeMeshGPT~\citep{lionar2025treemeshgpt}, and DeepMesh~\citep{zhao2025deepmesh}. 
It is worth noting that several strong methods have appeared recently; however, most do not release inference code or pre-trained weights.
Given the substantial cost of training mesh generation models, we restrict our comparisons to the four baselines with publicly available weights.
We also exclude closed-source commercial systems (e.g., Tripo 3D~\cite{tripo}), which typically employ substantially larger models and do not provide reproducible code.
For Hunyuan3D~\cite{lei2025hunyuan3dstudio}, which is a commercial closed-source system built upon scaled-up BPT~\cite{bpt}, we only compare against its open-source 0.5B variant, whose backbone size matches ours.

\paragraph{Indicators}
We evaluate triangle-mesh generation using four complementary metrics: Normal Consistency (NC), Chamfer Distance (CD), Hausdorff Distance (HD), and F-score (F1).
Specifically, NC measures normal consistency and reflects surface orientation and local geometric fidelity; CD quantifies the average bidirectional point-to-point deviation between reconstructed and reference surfaces; HD  captures the worst-case geometric error and is sensitive to outliers and fine-scale artifacts; and F1 summarizes precision recall trade-offs under a distance threshold, indicating overall surface coverage and completeness.
\FF{For all metrics, we uniformly sample 100K points from both the predicted mesh and the ground-truth mesh. CD and HD are computed from the bidirectional nearest-neighbor distances between these two point sets, taking the mean and maximum respectively. F1 is computed as the harmonic mean of precision and recall at a distance threshold of $0.003$.}
Together, these metrics jointly characterize both average and worst-case geometric accuracy, as well as perceptual surface quality.

We randomly selected 50 shapes from ShapeNet~\cite{chang2015shapenet} and 100 shapes each from Thingi10K~\cite{zhou2016thingi10k} and Objaverse~\cite{objaverseXL} to form our quantitative test sets\FF{, which are strictly excluded from our training data}. \FF{This 250-shape test set is used consistently across all quantitative evaluations, ablation studies, and user studies throughout the paper.} \FFF{Fig.~\ref{fig:testdataset} shows an overview of these 250 models in our test dataset.} \FF{Since autoregressive mesh generation can produce occasional non-manifold elements, we apply PyMeshLab~\cite{pymeshlab} as a lightweight post-processing step to all methods for fair evaluation, following MeshMosaic~\cite{xu2025meshmosaic}.}
We evaluated our method against four baselines, MeshAnythingV2~\citep{chen2024meshanythingv2}, BPT~\citep{bpt}, TreeMeshGPT~\citep{lionar2025treemeshgpt}, and DeepMesh~\citep{zhao2025deepmesh}, and report the NC, CD, HD, and F1 metrics in Table~\ref{tab:comp_all}.
Our method consistently outperforms the baselines across multiple metrics on all three datasets, highlighting its superior representational capacity and stronger alignment with the input shape.
We further provide qualitative visualizations with baseline methods in Fig.~\ref{fig:comp1}. Overall, our method produces more complete shapes, higher mesh quality, and more artist-like topology.

\begin{table}[t]
% \vspace{-2mm}
\caption{\textbf{User study with SOTA methods on triangle mesh generation.} \FF{Each score is the mean ranking-based score over all participants (range $[0, 3]$; 1st=3, 2nd=2, 3rd=1, others=0). }
}
\centering
\vspace{-3mm}
\label{tab:user_study_tri}
\resizebox{.85\linewidth}{!}{
\begin{tabular}{c|ccccc}
\toprule
         & MeshAthV2   & TreeMeshGPT  & BPT  & DeepMesh & Ours
\\ \midrule
Scores & 0.18  & 0.57  & \textbf{1.4}   & 1.17  & \underline{\textbf{2.61}} \\
\FF{Variance} & \FF{0.27} & \FF{0.67} & \FF{0.95} & \FF{0.93} & \FF{0.49} \\

\bottomrule
\end{tabular}
}
\end{table}

\paragraph{User Study.}
However, it is often difficult to quantitatively assess whether a generated mesh truly matches the characteristics of an artist-created mesh, as opposed to one produced by generic geometric processing. We therefore conduct a user study to evaluate how artist-like our generated meshes appear.
We recruited 25 professionals from the 3D industry as volunteers to conduct subjective evaluations.
\FF{Each participant evaluated 30 shape groups (10 for triangle mesh, 10 for quad mesh, and 10 for UV segmentation). For each group, participants were presented with rendered images from four viewpoints together with the ground-truth shape and the input point cloud. The four criteria (regularity, artist-likeness, geometric fidelity, and shape consistency) served as holistic guidelines; participants gave a single overall top-3 ranking rather than separate per-criterion scores. Rankings were converted to scores as 1st=3, 2nd=2, 3rd=1, and others=0.}
Table~\ref{tab:user_study_tri}
summarizes the comparative ratings from the user study. Overall, our method receives higher rankings from participants, indicating improved mesh quality and closer stylistic alignment with artist-created meshes.

\begin{figure*}[!tp]
    \centering
    \begin{overpic}[width=\linewidth]{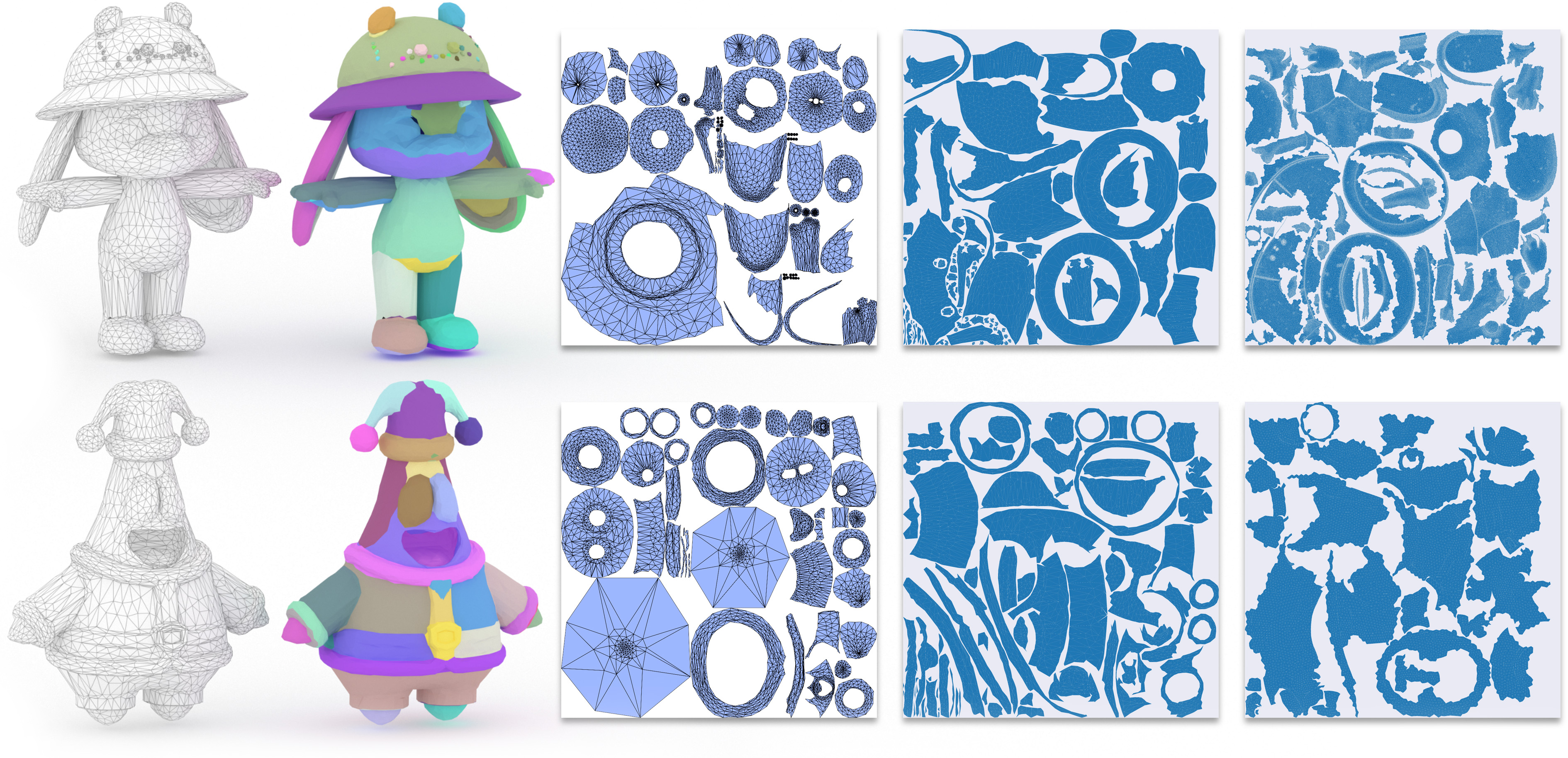}
    \put(1,  0.5){\textbf{(a) Generated Mesh}}
    \put(18, 0.5){\textbf{(b) Generated UV Seg}}
    \put(41, 0.5){\textbf{(c) Ours UV}}
    \put(57, 0.5){\textbf{(d) PartUV from Our Mesh}}
    \put(79, 0.5){\textbf{(e) PartUV from GT Mesh}}
    \end{overpic}
    \vspace{-8mm}
    \caption{\textbf{Qualitative comparison with PartUV~\cite{PartUV}.} Our method generates an artist mesh (a) together with explicit UV segmentation (b). 
By applying angle-based UV unwrapping from Blender~\cite{Blender}, we further obtain a high-quality 2D UV layout (c). 
In contrast, PartUV relies on a PartField~\cite{liu2025partfield} pre-segmentation pipeline; regardless of whether it is applied to our generated mesh (d) or the ground-truth (GT) mesh (e), its resulting UV charts are consistently less clean and less well-structured than ours. 
}
    \label{fig:UVcomp}
    \vspace{-3mm}
\end{figure*}

\begin{figure}[!tp]
    \centering
    \begin{overpic}[width=.8\linewidth]{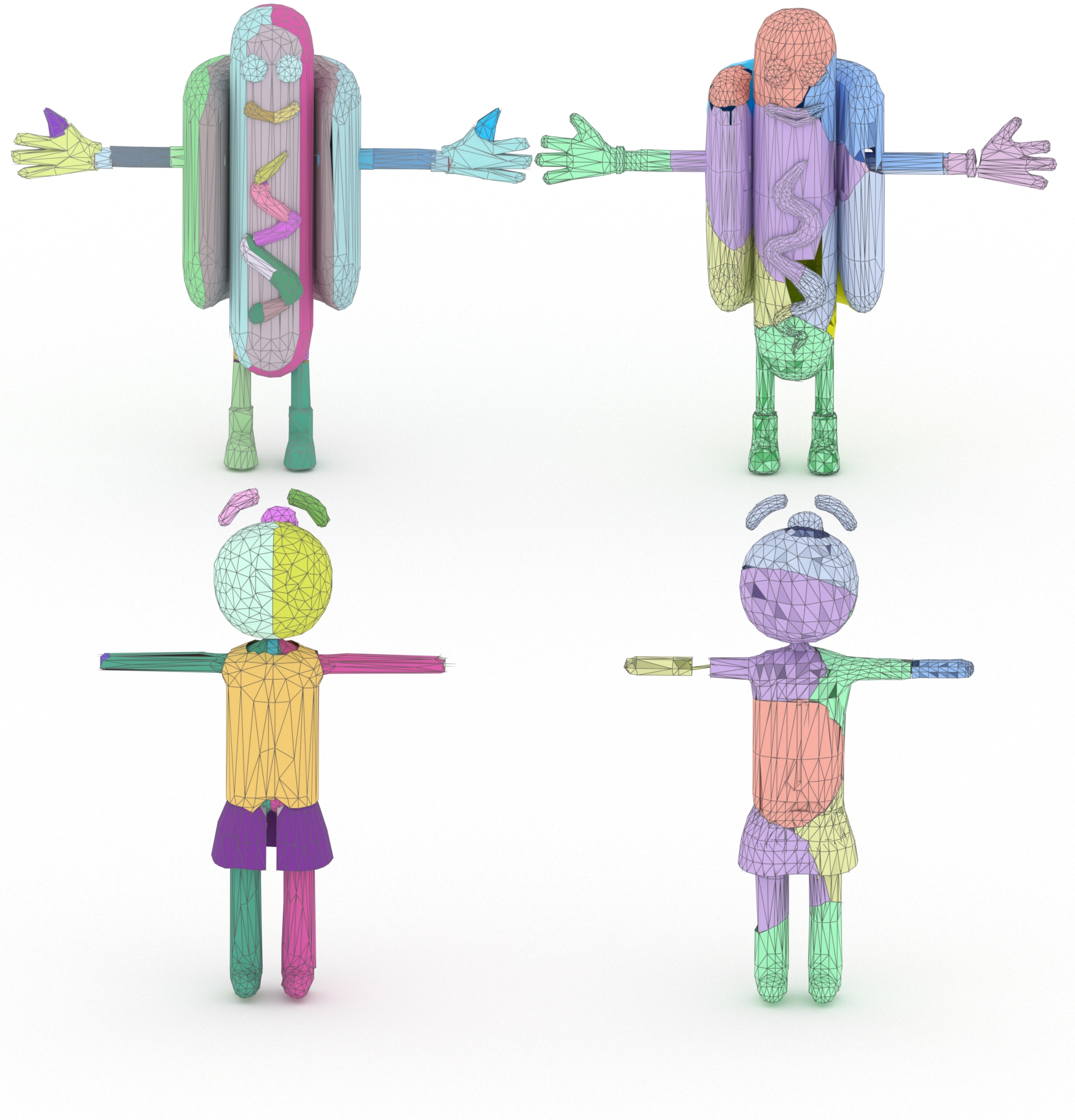}
    \put(20, 7){\textbf{Ours}}
    \put(60, 7){\textbf{MeshMosaic}}
    \end{overpic}
    \vspace{-8mm}
    \caption{\textbf{Comparison with MeshMosaic~\cite{xu2025meshmosaic}.} Our method yields cleaner, more regular segmentation and mitigates the issue of overly long seams.}
    \label{fig:mosaic}
    \vspace{-4mm}
\end{figure}

\begin{table}[t]
% \vspace{-2mm}
\caption{\textbf{User study with PartUV~\cite{PartUV}.} The scores are calculated based on the rankings and range from $[0, 3]$.}
\centering
\vspace{-4mm}
\label{tab:user_study_UV}
\resizebox{.85\linewidth}{!}{
\begin{tabular}{c|ccc}
\toprule
         & PartUV w/ Our Mesh   & PartUV w/ GT Mesh  & Ours  
\\ \midrule
Scores & \textbf{2.04} & 1.36 & \underline{\textbf{2.6}} \\
\FF{Variance} & \FF{0.49} & \FF{0.36} & \FF{0.38} \\

\bottomrule
\end{tabular}
}
\vspace{-3mm}
\end{table}

\FF{
\begin{table}[t]
\caption{\textbf{\FFF{UV distortion comparison.}} Our segmentation produces consistently lower UV distortion than PartField~\cite{liu2025partfield} across all four standard metrics with Blender's~\cite{Blender} angle-based unwrapping.}
\centering
\vspace{-3mm}
\label{tab:uv_distortion}
\resizebox{.95\columnwidth}{!}{
\begin{tabular}{c|cccc}
\toprule
Method & L2 Stretch ($\downarrow$ to 1) & Area Dist. ($\downarrow$) & Angle Dist. ($\downarrow$ to 1) & Sym. Dirichlet ($\downarrow$ to 4) \\
\midrule
PartField & 0.921 & 0.849 & 1.256 & 8.283 \\
SATO & \textbf{0.979} & \textbf{0.562} & \underline{\textbf{1.128}} & \textbf{5.156} \\
\FFF{Ground Truth} & \underline{\textbf{1.010}} & \underline{\textbf{0.146}} & \textbf{1.138} & \underline{\textbf{4.102}} \\
\bottomrule
\end{tabular}
}
\end{table}
}

\subsection{UV Segmentation}
\label{sec:UVcomparison}
Simultaneously generating UV segmentation during autoregressive mesh synthesis remains largely unexplored. To the best of our knowledge, \name is the first method to explicitly support this task, which makes direct comparisons challenging. The closest recent open-source baseline is PartUV~\cite{PartUV}, which performs UV segmentation on an input mesh with PartField~\cite{liu2025partfield} segmentation but does not generate meshes and instead operates on the provided geometry.

Fig.~\ref{fig:UVcomp} reports a qualitative comparison with PartUV~\cite{PartUV}, where (a, b) show the triangular mesh and UV segmentation result produced by our method, and (c) visualizes the UV layout obtained by unwrapping our predicted segmentation in Blender~\cite{Blender}. In contrast, Fig.~\ref{fig:UVcomp} (d,e) present PartUV’s Blender unwrappings when applied to (d) our generated triangular mesh and (e) a high-quality ground-truth triangular mesh, respectively. 
Our method yields consistently cleaner and higher-quality UV layouts, whereas PartUV~\cite{PartUV} produces less regular unwrappings regardless of whether it is applied to our generated mesh or the ground-truth mesh.
Furthermore, when PartUV~\cite{PartUV} is applied to our generated high-quality triangular meshes, it produces better UV unwrapping results than when applied to the original GT meshes. This also highlights the practicality and downstream usability of our generated triangular meshes.
We further present a gallery of our UV unwrapping results in Fig.~\ref{fig:UVgallery}, where the model is derived from Fig.~\ref{fig:gallery} and unwrapped using Blender~\cite{Blender} following PartUV~\cite{PartUV}.

We also conduct a user study for the UV unwrapping evaluation. Participants were asked to rate the UV layouts produced by our method and by PartUV~\cite{PartUV}. Table~\ref{tab:user_study_UV} reports the quantitative results, showing that artists consistently prefer the cleaner, more organized UV segmentation achieved by our approach.

\FF{
\paragraph{UV Distortion.}
To quantitatively evaluate the UV quality resulting from our segmentation, we compute standard parameterization distortion metrics on the 10 generated meshes from Fig.~\ref{fig:UVgallery}.
For each mesh, we apply Blender's angle-based unwrapping~\cite{Blender}, then compute the Jacobian of the 3D-to-UV mapping per triangle, obtain its singular values $\sigma_1 \geq \sigma_2 > 0$, and report four metrics: L2 Stretch $\sqrt{(\sigma_1^2+\sigma_2^2)/2}$, Area Distortion $|\log(\sigma_1 \cdot \sigma_2)|$, Angle Distortion $\sigma_1/\sigma_2$, and Symmetric Dirichlet energy $\sigma_1^2+\sigma_2^2+1/\sigma_1^2+1/\sigma_2^2$~\cite{smith2015bijective}.
All values are area-weighted medians after per-island normalization.
As the baseline, we compare against PartField~\cite{liu2025partfield} segmentation unwrapped with the same algorithm.
Table~\ref{tab:uv_distortion} shows that our segmentation consistently yields lower distortion across all four metrics, indicating that our predicted chart boundaries better align with geometric features, producing more regular islands that are easier to unwrap with low distortion.
\FFF{Even compared with ground truth UVs created by artists, our method still shows advantages in certain aspects, such as angle distortion.}
}

Another autoregressive mesh generation method related to segmentation is a very recent work, MeshMosaic~\cite{xu2025meshmosaic}.
It leverages PartField~\cite{liu2025partfield} predicted segmentations to decompose a shape into multiple parts and then autoregressively generates the mesh part by part in a fixed sequence. Fig.~\ref{fig:mosaic} compares MeshMosaic~\cite{xu2025meshmosaic} with our method. While both approaches can follow the input shape, MeshMosaic’s reliance on precomputed part boundaries often yields unnatural transitions across parts, including visible seams, and can introduce asymmetry artifacts. In contrast, our method jointly generates the full mesh and its segmentation in a unified pass, which naturally enforces global consistency and avoids inter-part seams and symmetry issues.

\begin{figure}[!tp]
    \centering
    \begin{overpic}[width=.9\linewidth]{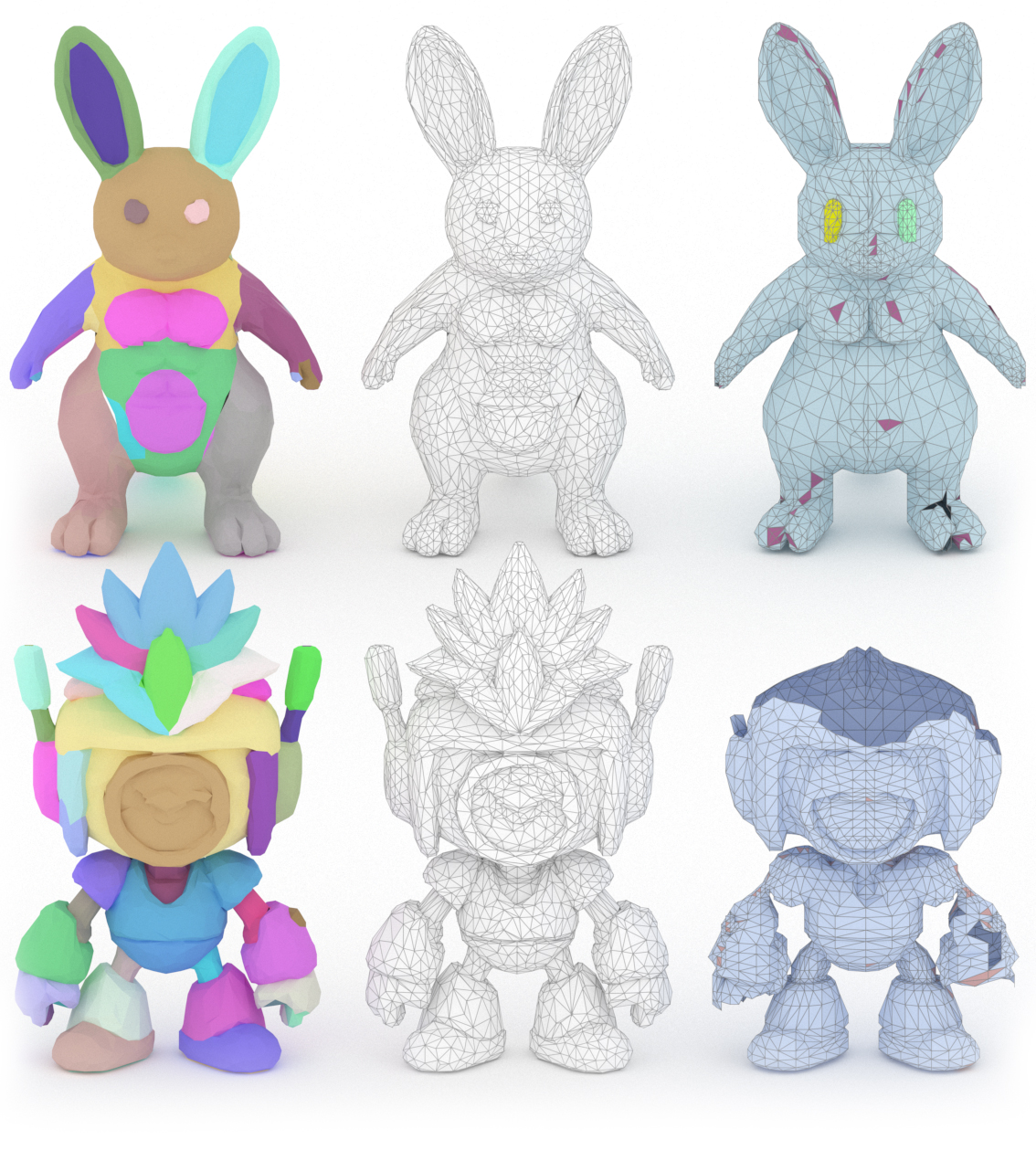}
    \put(8, 3){\textbf{Ours Seg}}
    \put(37, 3){\textbf{Ours Mesh}}
    \put(65, 3){\textbf{MeshSilksong}}
    \end{overpic}
    \vspace{-6mm}
    \caption{\textbf{Comparison with MeshSilksong~\cite{song2025meshSilksong}.} Our method produces higher-quality meshes and cleaner, more coherent segmentation.}
    \label{fig:silk}
    \vspace{-2mm}
\end{figure}

More recently, MeshSilksong~\cite{song2025meshSilksong} can predict connected components during autoregressive mesh generation. Although these labels are not UV segmentations, we include MeshSilksong~\cite{song2025meshSilksong} as an additional point of comparison. Fig.~\ref{fig:silk} visualizes results from our method and MeshSilksong, where MeshSilksong uses different colors to denote different connected components, whereas our colors indicate UV charts. 
MeshSilksong almost failed at the segmentation task, only separating the rabbit's eyes, while the rest of the whole was treated as a single connected component as output.
Overall, our method produces meshes that are more complete and higher quality and yields more coherent, meaningful segmentations, highlighting the advantage of our joint generation approach.

\begin{figure}[!tp]
    \centering
    \begin{overpic}[width=\linewidth]{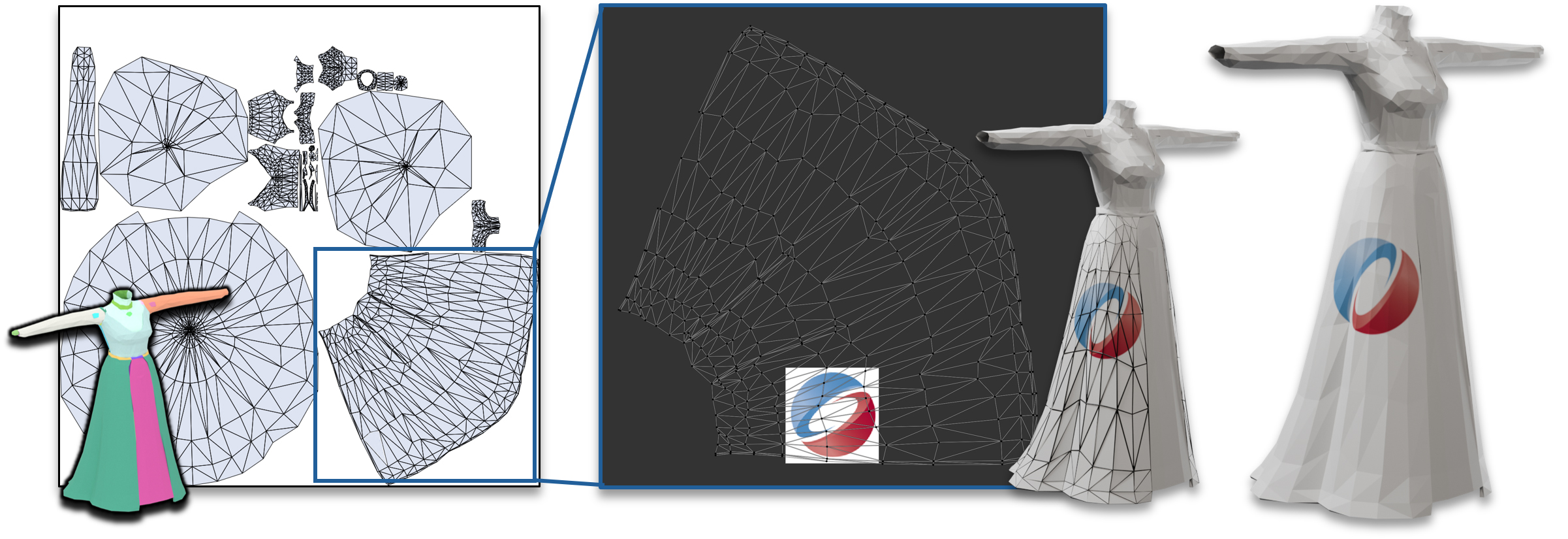}
    \end{overpic}
    \vspace{-8mm}
    \caption{\textbf{Texture painting with our UV unwrapping.} The high-quality UV unwrapping produced by our method makes it easy for artists to paint texture maps.}
    \label{fig:app_uv}
    \vspace{-4mm}
\end{figure}

Finally, we demonstrate the practical utility of the UV unwrapping produced by our method in Fig.~\ref{fig:app_uv}. Artists can readily apply textures to the resulting UV layout: each component of the input shape is cleanly and consistently separated into well-defined islands, enabling targeted texture painting without inadvertently affecting other parts.

\begin{figure*}[!tp]
    \centering
    \begin{overpic}[width=.98\linewidth]{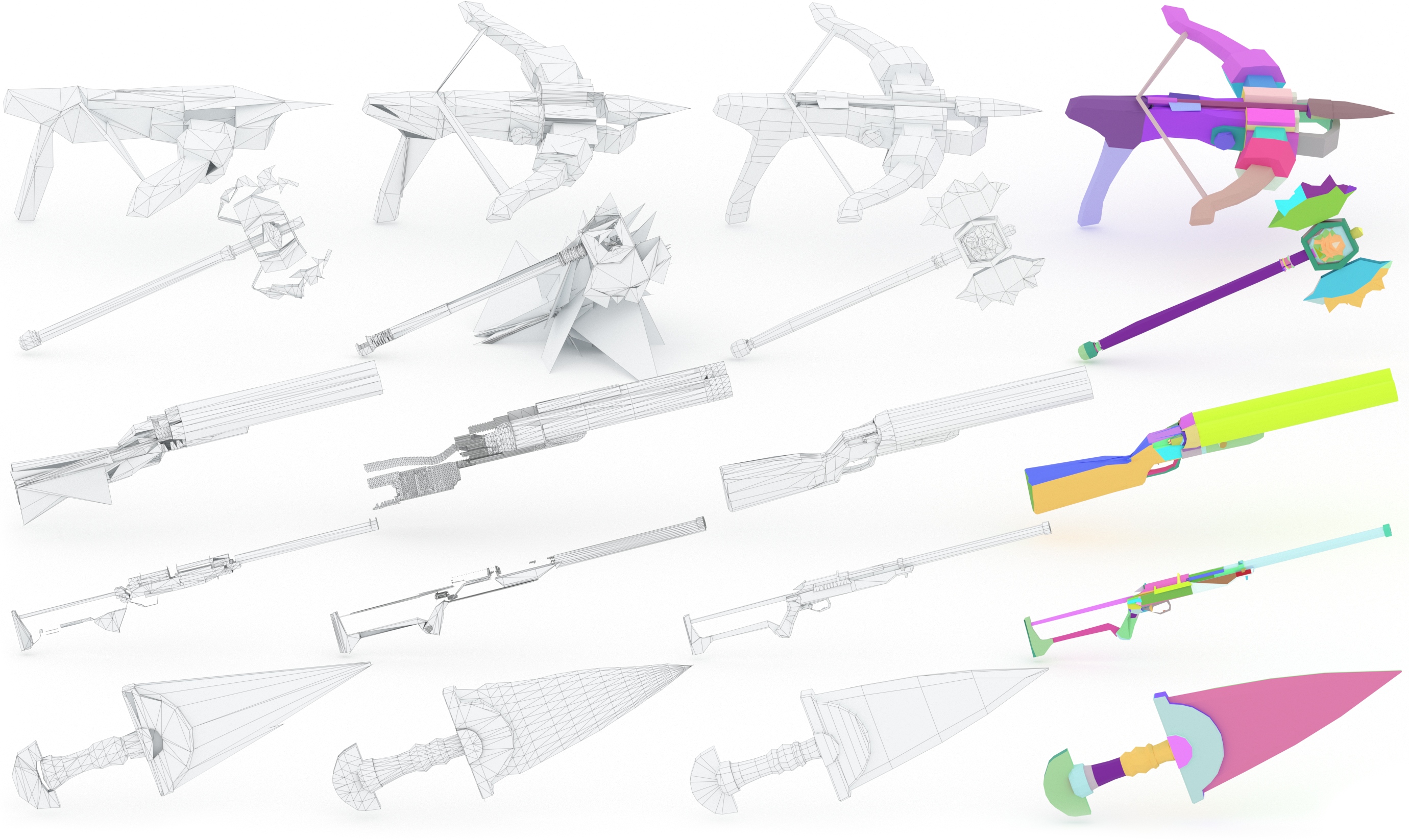}
    \put(10, 1){\textbf{BPT}}
    \put(30, 1){\textbf{DeepMesh}}
    \put(57, 1){\textbf{Ours}}
    \put(80, 1){\textbf{Ours UV}}
    \end{overpic}
    \vspace{-5mm}
    \caption{\textbf{\FF{Qualitative comparison with BPT~\cite{bpt} and DeepMesh~\cite{zhao2025deepmesh} on diverse shapes.}}
Compared with prior triangle-mesh generation models, our method more consistently generates high-quality quadrilateral meshes, is more stable, and additionally predicts native UV segmentation.}
    \label{fig:quadcomp}
    \vspace{-4mm}
\end{figure*}

\begin{figure}[!tp]
    \centering
    % \vspace{-1mm}
    \begin{overpic}[width=\linewidth]{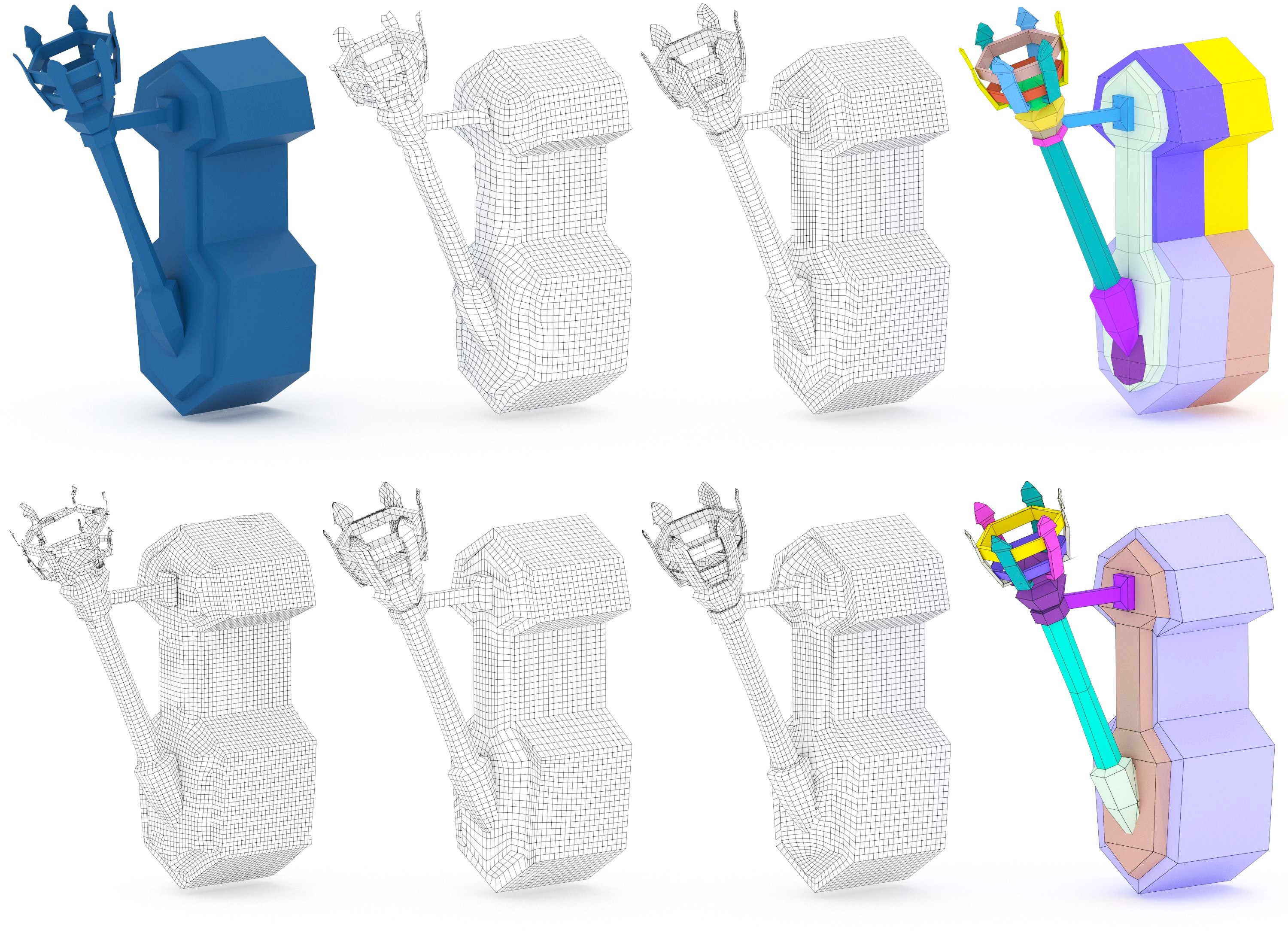}
    \put(11, 38){\textbf{Input}}
    \put(31, 38){\textbf{Quadriflow}}
    \put(59, 38){\textbf{Quadwild}}
    \put(85, 38){\textbf{Ours}}
    \put(14,  1){\textbf{IM}}
    \put(31,  1){\textbf{NeurCross}}
    \put(59, 1){\textbf{CrossGen}}
    \put(85, 1){\textbf{Ours}}
    \end{overpic}
    \vspace{-8mm}
    \caption{\FFF{\textbf{Qualitative comparison with quad remeshing and reconstruction methods.}}
Due to reliance on quadrilateral parameterization, these methods typically struggle to produce highly simplified quad meshes. In contrast, our method can generate meshes \FFF{with diverse densities}  and additionally supports native UV segmentation.}
    \label{fig:quadcomp1}
    \vspace{-2mm}
\end{figure}
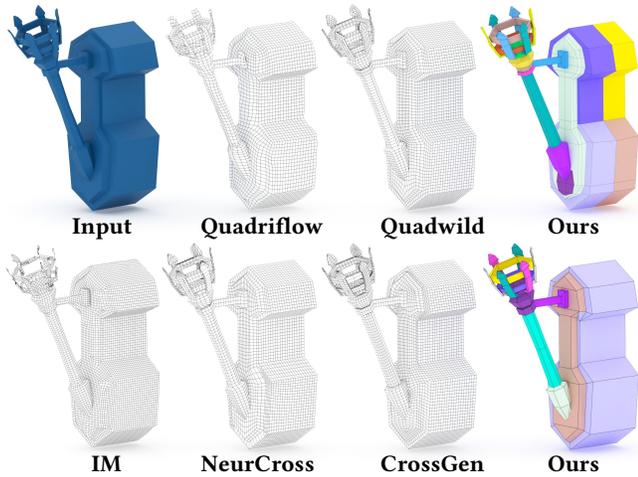

\subsection{Quad Mesh Generation}
\label{sec:Quadcomparison}

Finally, we evaluate the quadrilateral meshes generated by our method. As described in Sec.~\ref{sec:decoding}, \name supports quad-mesh generation with a simple switch of the detokenizer, without altering the model architecture.
\FF{Since our quad detokenizer simply merges adjacent triangle pairs from the same token sequence, the geometric fidelity of quad outputs is nearly identical to that of the corresponding triangle outputs.}

For assessing quad quality, the recently released QuadGPT~\cite{QuadGPT} is a strong reference baseline; however, it does not publicly provide code or pretrained weights\FF{, and is trained on an undisclosed proprietary dataset of 1.3M quad meshes, making direct comparison infeasible}. \FF{Following QuadGPT, we do not enforce strict quad coplanarity, as artist-created quad meshes in practice also exhibit slight non-planarity.} We therefore follow QuadGPT’s evaluation protocol and compare against representative triangle-mesh autoregressive methods. Fig.~\ref{fig:quadcomp} presents a visual comparison between the quadrilateral meshes produced by our method and those obtained by BPT~\cite{bpt} and DeepMesh~\cite{zhao2025deepmesh}. Our approach not only yields high-quality, high-fidelity quad meshes, but also simultaneously produces clean, artist-aligned UV segmentations.

To further demonstrate both the capability and practical value of our quad-mesh generator, we additionally compare against several established quadrilateral reconstruction and remeshing methods. Fig.~\ref{fig:quadcomp1} contrasts our results with five alternatives: IM~\cite{IM_Instant_Meshes2015}, QuadriFlow~\cite{QuadriFlow2018}, QuadWild~\cite{quadwild2021}, NeurCross~\cite{Dong2025NeurCross}, and CrossGen~\cite{Dong2025CrossGen}. IM, QuadriFlow, and QuadWild are classical parameterization-based quad remeshing approaches, whereas NeurCross and CrossGen represent more recent cross-field generation methods.
\FF{All baselines were run using their default parameter settings. To produce outputs of comparable resolution across methods while preserving sufficient geometric detail, we set the corresponding resolution-control parameters for each baseline, namely, 3000 points for IM, 6000 faces for QuadriFlow, and a scaleFact parameter of 1.0 for QuadWild, NeurCross, and CrossGen. Because these methods generate isotropic quadrilateral meshes, reducing the resolution leads to a simpler output but inevitably sacrifices geometric detail. }
As shown in Fig.~\ref{fig:quadcomp1}, these remeshing baselines often struggle to simultaneously achieve high quad utilization, low face count, and consistent alignment with salient feature lines. In contrast, our method produces compact, well-structured quad layouts that better match artist expectations. This comparison not only highlights the quality of our outputs, but also underscores the importance of autoregressive artist mesh generation as a complementary direction to conventional remeshing pipelines.
\FF{We additionally report geometric metrics for the shape in Fig.~\ref{fig:quadcomp1} in Table~\ref{tab:quad_geo}. Remeshing methods achieve near-identical fidelity since they operate directly on the ground-truth geometry, whereas our method generates the mesh from a point cloud; despite this, our output achieves competitive or superior scores.}
\begin{table}[t]
% \vspace{-2mm}
\caption{\textbf{User study with remeshing and reconstruction methods on quad mesh generation.} The scores are calculated based on the rankings and range from $[0, 3]$.
}
\centering
\vspace{-4mm}
\label{tab:user_study_quad}
\resizebox{.95\linewidth}{!}{
\begin{tabular}{c|cccccc}
\toprule
         & IM   & QuadriFlow  & Quadwild  & NeurCross & CrossGen  & Ours
\\ \midrule
Scores & 0.12 & 0.28 &  1.08 & \textbf{1.48}  & 1.24 & \underline{\textbf{1.8}} \\
\FF{Variance} & \FF{0.20} & \FF{0.44} & \FF{1.23} & \FF{1.32} & \FF{1.28} & \FF{1.28} \\

\bottomrule
\end{tabular}
}
\vspace{-2mm}
\end{table}
\FF{
\begin{table}[t]
\caption{\textbf{\FF{Geometric metrics on the quad mesh from Fig.~\ref{fig:quadcomp1}.}} Remeshing methods achieve near-identical fidelity to the input since they operate directly on the ground-truth geometry, whereas our method generates the mesh from a point cloud.}
\centering
\vspace{-3mm}
\label{tab:quad_geo}
\resizebox{.6\columnwidth}{!}{
\begin{tabular}{c|cccc}
\toprule
Method & $\mathrm{NC}\uparrow$ & $\mathrm{CD}\downarrow$ & $\mathrm{HD}\downarrow$ & $\mathrm{F1}\uparrow$ \\
\midrule
IM          & 0.917 & 0.007 & \textbf{0.052} & 0.304 \\
QuadriFlow  & 0.924 & 0.005 & 0.080 & 0.451 \\
QuadWild    & \textbf{0.970} & \underline{\textbf{0.001}} & \underline{\textbf{0.020}} & 0.848 \\
NeurCross   & 0.968 & \textbf{0.002} & \underline{\textbf{0.020}} & 0.846 \\
CrossGen    & 0.969 & \underline{\textbf{0.001}} & \underline{\textbf{0.020}} & \textbf{0.849} \\
\textbf{Ours} & \underline{\textbf{0.971}} & \underline{\textbf{0.001}} & \underline{\textbf{0.020}} & \underline{\textbf{0.857}} \\
\bottomrule
\end{tabular}
}
\vspace{-2mm}
\end{table}
}
\begin{table}[!tp]
\centering
\vspace{1mm}
\caption{\textbf{Quantitative comparison with the DeepMesh~\cite{zhao2025deepmesh} tokenizer.}
Under the overfitting setup in Fig.~\ref{fig:abla}, we report compression and training metrics for both tokenizers. Our method achieves better training speed and a higher compression ratio.}
\vspace{-4mm}
\resizebox{\columnwidth}{!}{
    \begin{tabular}{c|ccccc}
    \toprule
Methods              & Token Length $\downarrow$ & Transitions $\downarrow$ & Time (s) $\downarrow$ & Training Speed (Steps/s) $\uparrow$ & \FF{Infer (Tokens/s) $\uparrow$} \\ \midrule

DeepMesh             & 24674      & 1654        & 2.061 & 0.442 & \FF{$\sim$55} \\
Ours                 & \underline{\textbf{20830}}      & \underline{\textbf{981}}         & \underline{\textbf{0.319}} & \underline{\textbf{0.488}} & \FF{\underline{\textbf{$\sim$58}}} \\
\bottomrule
\end{tabular}
}
\label{tab:compDMTOKEN}
\vspace{-1mm}
\end{table}

Furthermore, we also conduct a user study for this task, asking participants to evaluate the quad meshes produced by our method and by the five quadrilateral reconstruction/remeshing baselines described above. 
Table~\ref{tab:user_study_quad} summarizes the results, showing a clear preference for our outputs. This further validates the practicality of our artist quadrilateral mesh generation and highlights its value for real-world content creation workflows.

\subsection{Ablation Studies}
\label{sec:abla}
We conduct a series of ablation studies to verify our proposed ideas and to compare them in detail with other methods.

\subsubsection{Tokenizer}
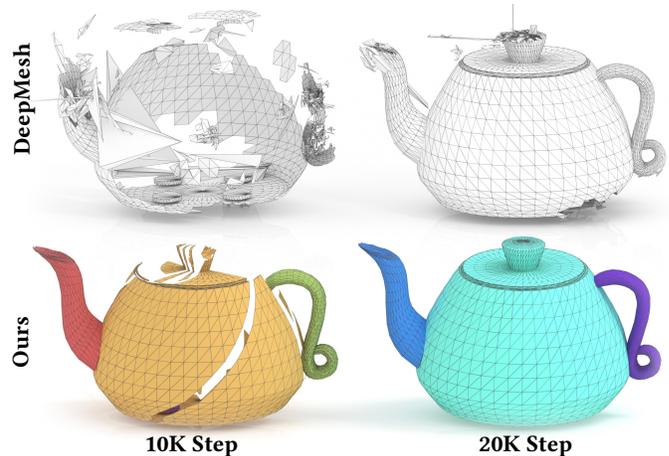
\begin{figure}[!tp]
    \centering
    \begin{overpic}[width=\linewidth]{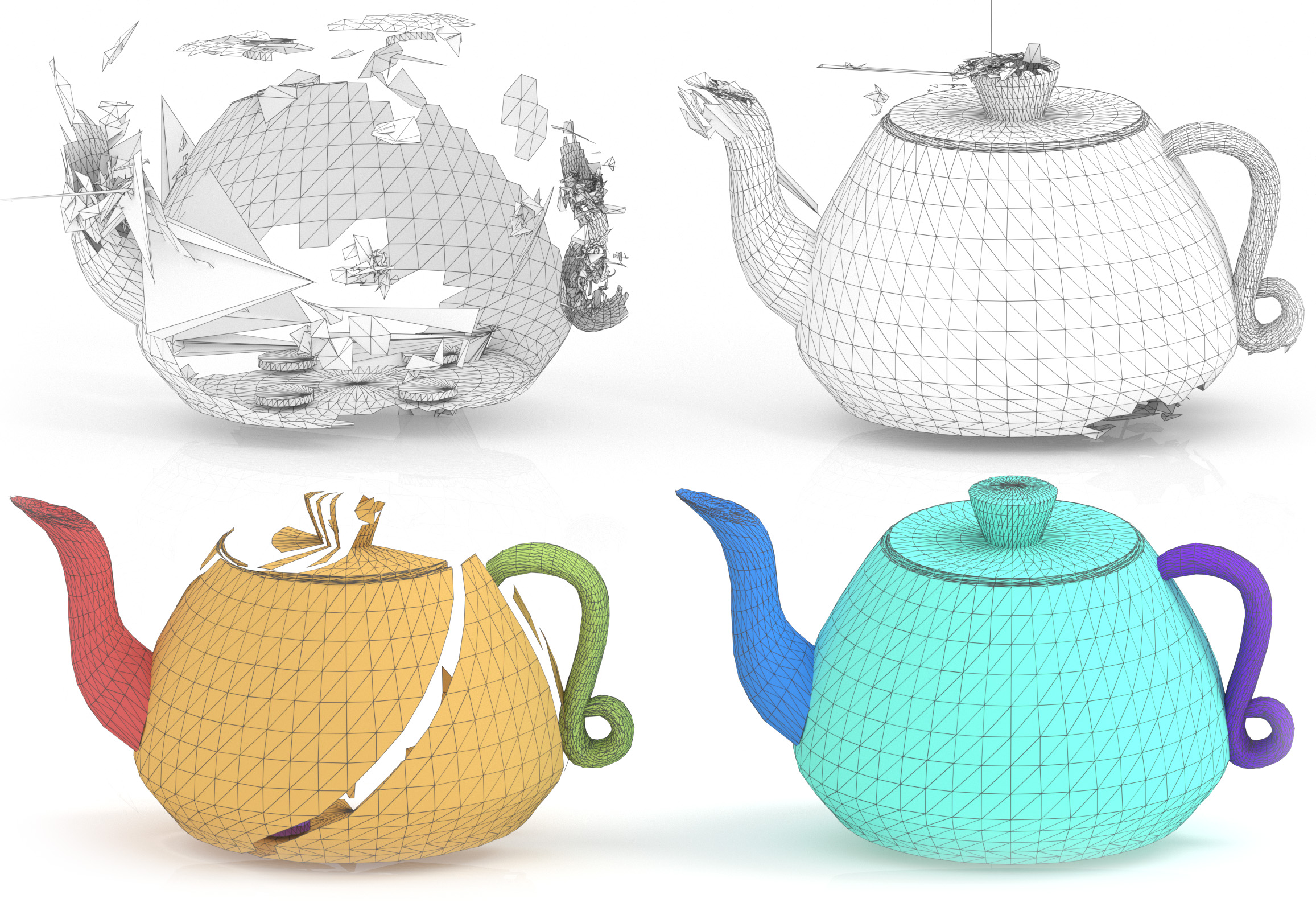}
    \put(-2.5, 46){\rotatebox{90}{\textbf{DeepMesh}}}
    \put(-2.5, 13 ){\rotatebox{90}{\textbf{Ours}}}
    \put(18, 0){\textbf{10K Step}}
    \put(70, 0){\textbf{20K Step}}
    \end{overpic}
    \vspace{-7mm}
    \caption{\textbf{Ablation with the DeepMesh~\cite{zhao2025deepmesh} tokenizer.}
We constructed an overfitting ablation to compare our tokenizer with DeepMesh. Our tokenizer converges faster and is easier for the network to learn, even when augmented with UV segmentation.}
    \label{fig:abla}
    \vspace{-4mm}
\end{figure}

First, we validate the advantage of \name via a more fine-grained comparison with DeepMesh’s tokenizer~\cite{zhao2025deepmesh}. Specifically, we construct an overfitting experiment on the teapot model in Fig.~\ref{fig:abla}, training with either \name or DeepMesh tokenization. To prevent the network from trivially memorizing a fixed token sequence, we apply random rotations to the input shape during training.

We use the same 0.5B Hourglass Transformer architecture and identical training hyperparameters for both settings: training on 8$\times$A800 GPUs for 20{,}000 steps, with the tokenizer as the only difference. Fig.~\ref{fig:abla} visualizes the generation results at 10{,}000 and 20{,}000 steps. Our method learns UV segmentation cues and reaches a near-perfect reconstruction noticeably faster. Even at early stages, the predicted segmentation is already clean and well-structured. Moreover, thanks to our strip-based serialization, the intermediate geometry appears cleaner and more refined, indicating easier optimization and faster convergence.

We also record a set of training statistics for this overfitting experiment, summarized in Table~\ref{tab:compDMTOKEN}. When encoding the teapot model, the two tokenizers produce sequences of 24K and 20K tokens, respectively, meaning \name requires only about 85\% of the token length of DeepMesh. This reduction is largely attributable to fewer patch/strip transitions (0.9K vs. 1.6K). Moreover, thanks to our efficient next-face query structure in the tokenization code, \name substantially reduces encoding time relative to the baseline, which in turn translates into a clear advantage in overall training throughput.

\FF{
To further isolate the benefit of our tokenizer at scale, we conduct a controlled large-scale comparison.
Since DeepMesh~\cite{zhao2025deepmesh} only releases inference code and pretrained weights but not its training pipeline, we re-implement its tokenizer within our training framework so that both settings share the identical model architecture, training data, optimizer, and hyperparameters.
Both models are trained on $64\times$A800 GPUs for 200K steps, with the tokenizer as the sole variable.
Table~\ref{tab:abla_largescale} reports the results on our 250-shape test set, confirming that the strip-based tokenizer yields consistent improvements across all metrics under strictly matched training conditions.
}

\FF{
\begin{table}[t]
\caption{\textbf{\FF{Large-scale tokenizer ablation.}} Both models use the same architecture, data, and training budget ($64\times$A800, 200K steps); only the tokenizer differs.}
\centering
\vspace{-3mm}
\label{tab:abla_largescale}
\resizebox{.65\columnwidth}{!}{
\begin{tabular}{c|cccc}
\toprule
Tokenizer & $\mathrm{NC}\uparrow$ & $\mathrm{CD}\downarrow$ & $\mathrm{HD}\downarrow$ & $\mathrm{F1}\uparrow$ \\
\midrule
DeepMesh & 0.908 & 0.022 & 0.108 & 0.455 \\
Ours & \underline{\textbf{0.925}} & \underline{\textbf{0.014}} & \underline{\textbf{0.103}} & \underline{\textbf{0.560}} \\
\bottomrule
\end{tabular}
}
\vspace{-2mm}
\end{table}
}

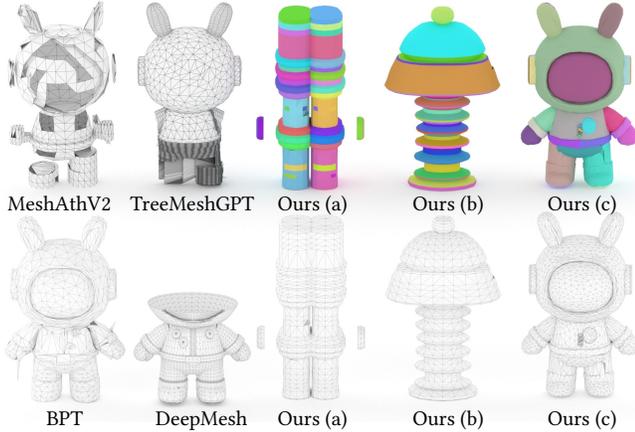
\begin{figure}[!tp]
    \centering
    \begin{overpic}[width=\linewidth]{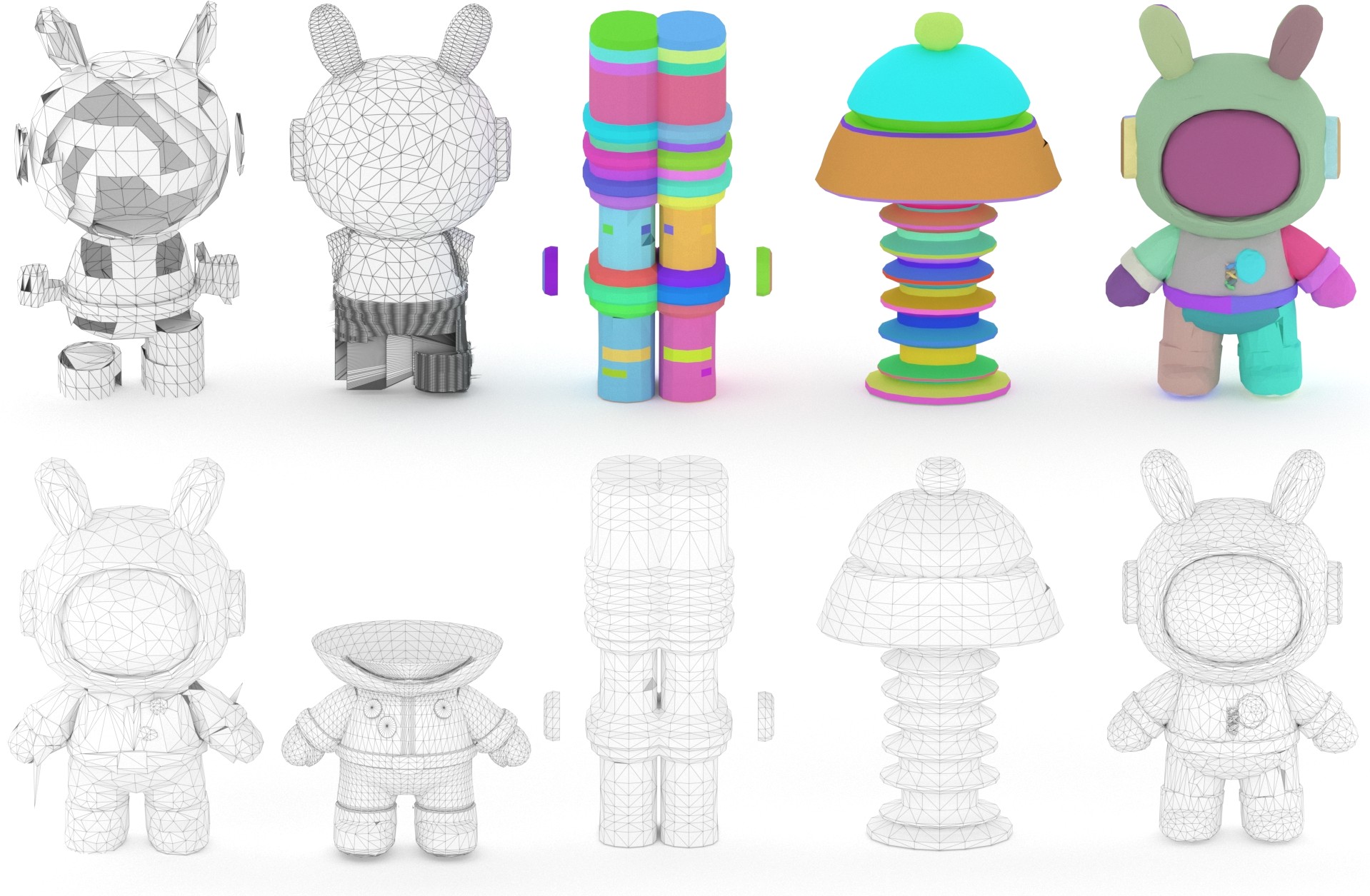}
    \put(1, 33){\small MeshAthV2}
    \put(20, 33){\small TreeMeshGPT}
    \put(7,  -0.5){\small BPT}
    \put(24, -0.5){\small DeepMesh}
    \put(43, 33){\small Ours (a)}
    \put(64, 33){\small Ours (b)}
    \put(85, 33){\small Ours (c)}
    \put(43, -0.5){\small Ours (a)}
    \put(64, -0.5){\small Ours (b)}
    \put(85, -0.5){\small Ours (c)}
    \end{overpic}
    \vspace{-7mm}
    \caption{\textbf{\FFF{Ablation on UV training strategies with a manually generated unseen test shape.}}
\FFF{All methods use the same test shape, which is manually created with Rodin~\cite{CLAY} from an input text prompt \textit{cute bunny astronaut toy}, converted into an SDF, and then meshed with marching cubes. This construction guarantees that the shape lies strictly outside the training set.
We compare SOTA methods and three training strategies: (a) using a pretrained point cloud VAE with UV segmentation data, (b) training from scratch on UV data, and (c) first training from scratch without UV data followed by post training on UV data. The first two strategies that directly use UV data cannot reliably align with the input shape and usually collapse to an almost random shape.}}
    \label{fig:uvabla}
\end{figure}

\subsubsection{Pretraining for UV}
\label{sec:abla_uv_pretrain}
\FFF{As discussed in Sec.~\ref{sec:train}, we first pre-train our network on triangle mesh data without UV segmentation, and then post-train it on data with UV segmentation.
We find that imposing UV supervision from the beginning often prevents the model from learning fine grained geometric details and makes it more difficult to align the generated mesh with the input.

Fig.~\ref{fig:uvabla} compares different training strategies on a manually generated test shape that is guaranteed to be unseen during training.
Specifically, the input shape is manually created with Rodin~\cite{CLAY} from an input prompt \textit{cute bunny astronaut toy}, converted into an SDF, and then meshed with marching cubes, which ensures that it lies strictly outside the training set. }
All three settings use the same conditioning point-cloud input and are trained for three days on 256$\times$A800 GPUs. When trained entirely from scratch (\textit{From scratch} in Fig.~\ref{fig:uvabla}), the model collapses to producing essentially random shapes, with only a coarse alignment in orientation. Even replacing our point-cloud encoder with a pretrained, frozen Hunyuan 3D~\cite{lei2025hunyuan3dstudio} VAE encoder does not substantially alleviate this issue (\textit{Pretrained} in Fig.~\ref{fig:uvabla}). In contrast, our two-stage strategy that involves pre-training without UV followed by post-training with UV achieves accurate alignment with the input while producing clean, well-structured UV segmentations.

\begin{figure}[!tp]
    \centering
    \begin{overpic}[width=.7\linewidth]{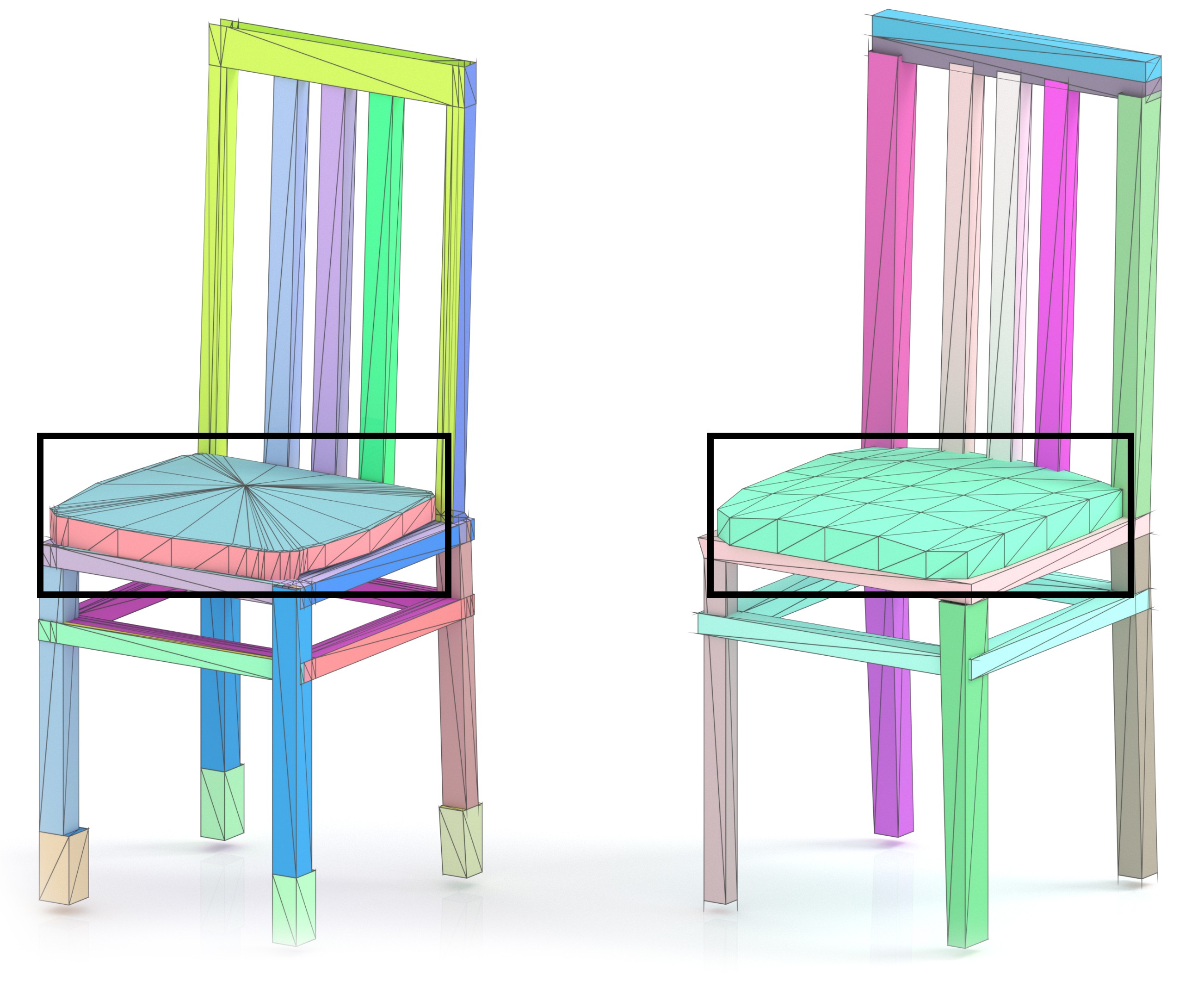}
    \put(6, -1){\textbf{w/o fine-tune}}
    \put(64, -1){\textbf{w/ fine-tune}}
    \end{overpic}
    \vspace{-2mm}
    \caption{\textbf{Ablation on quad-mesh fine-tuning.}
After quad-mesh fine-tuning, the meshes in the black boxed region become markedly higher quality and more artist-aligned, with cleaner structure and easier downstream editing.}
    \label{fig:quadabla}
    \vspace{-4mm}
\end{figure}

\subsubsection{Quad Mesh Fine-tuning}
\label{sec:quadabla}
Also discussed in Sec.~\ref{sec:train}, incorporating high-quality quadrilateral mesh data can further improve our triangle-mesh generator. In practice, fine-tuning on quad meshes encourages neater mesh routing and increases the prevalence of well-shaped triangles (often closer to right-angled triangles), bringing the output closer to artist-created meshes. Fig.~\ref{fig:quadabla} compares results before and after fine-tuning with quadrilateral data. After fine-tuning, the generated meshes exhibit cleaner, more quad-like routing and a reduced tendency to produce dense regions of elongated, skinny triangles.

\begin{figure}[!tp]
    \centering
    \begin{overpic}[width=\linewidth]{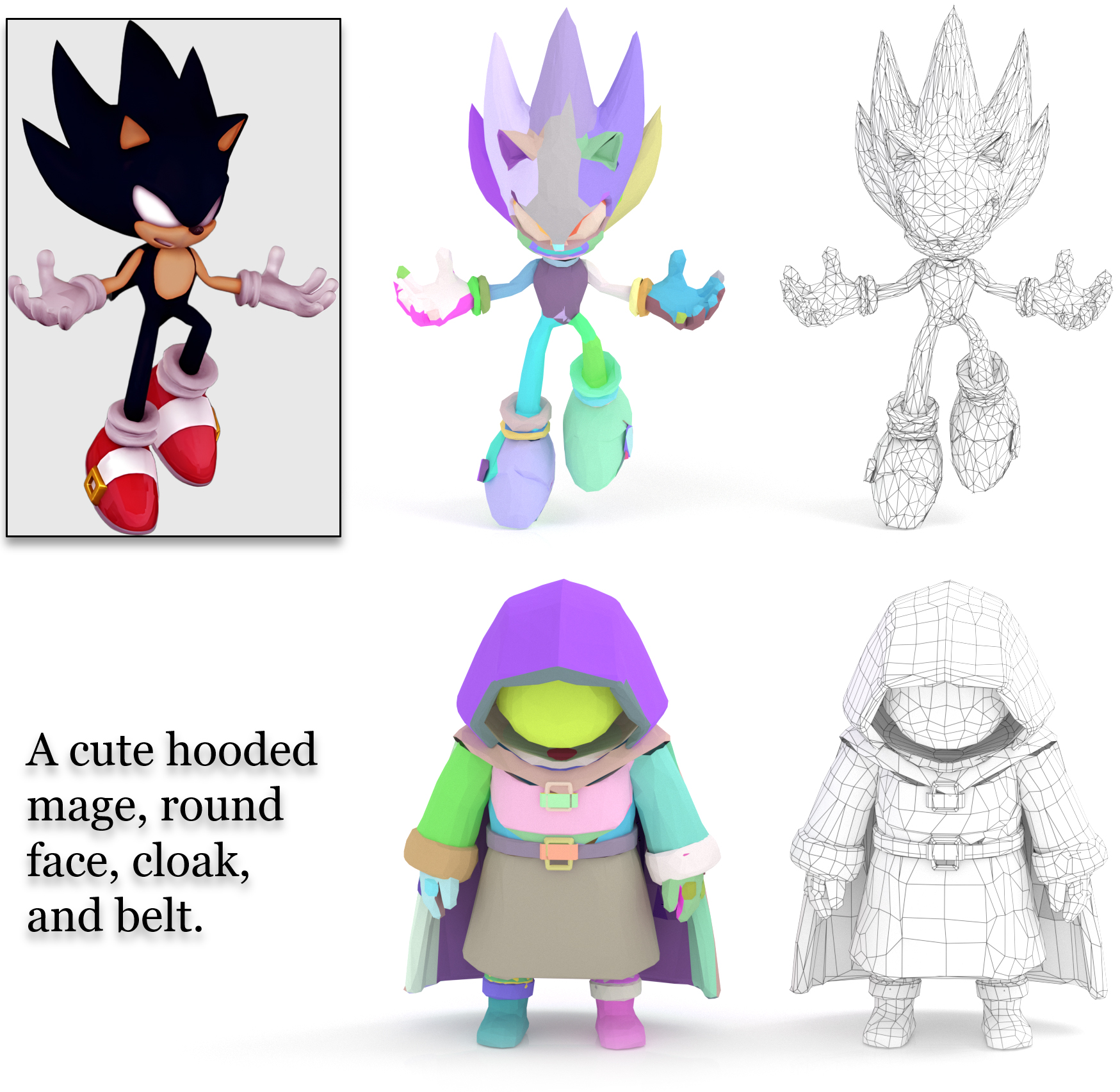}
    \end{overpic}
    \vspace{-10mm}
    \caption{\textbf{Generation from image and text prompts.}
By leveraging CLAY~\cite{CLAY} for 3D generation, \name can produce high-quality artist meshes with native UV segmentation from either an input image or a text prompt.}
    \label{fig:imagetext}
    \vspace{-4mm}
\end{figure}

\subsection{More Discussions}
\label{sec:discuss}

\paragraph{Generation with Image and Text Prompt.}
Techniques for generating 3D assets from diverse inputs have advanced rapidly in recent years~\cite{CLAY,lai2025hunyuan3d}. 
Many SDF-based pipelines can produce highly detailed geometry, but they typically yield ultra-dense triangle meshes via Marching Cubes~\cite{MarchingCubes}, and still require substantial post-processing or downstream meshing to obtain lightweight, production-ready, artist-style meshes. Fig.~\ref{fig:imagetext} demonstrates how our method can be used as a remeshing stage for such generated shapes.
Specifically, we use CLAY~\cite{CLAY} as an upstream generator; given an image or a text prompt, it predicts a 3D SDF and extracts a high-face-count mesh using Marching Cubes. Starting from this mesh, our method further performs UV segmentation and produces high-quality, lightweight triangular or quadrilateral meshes that are directly usable in practice.

\begin{figure}[!tp]
    \centering
    \begin{overpic}[width=\linewidth]{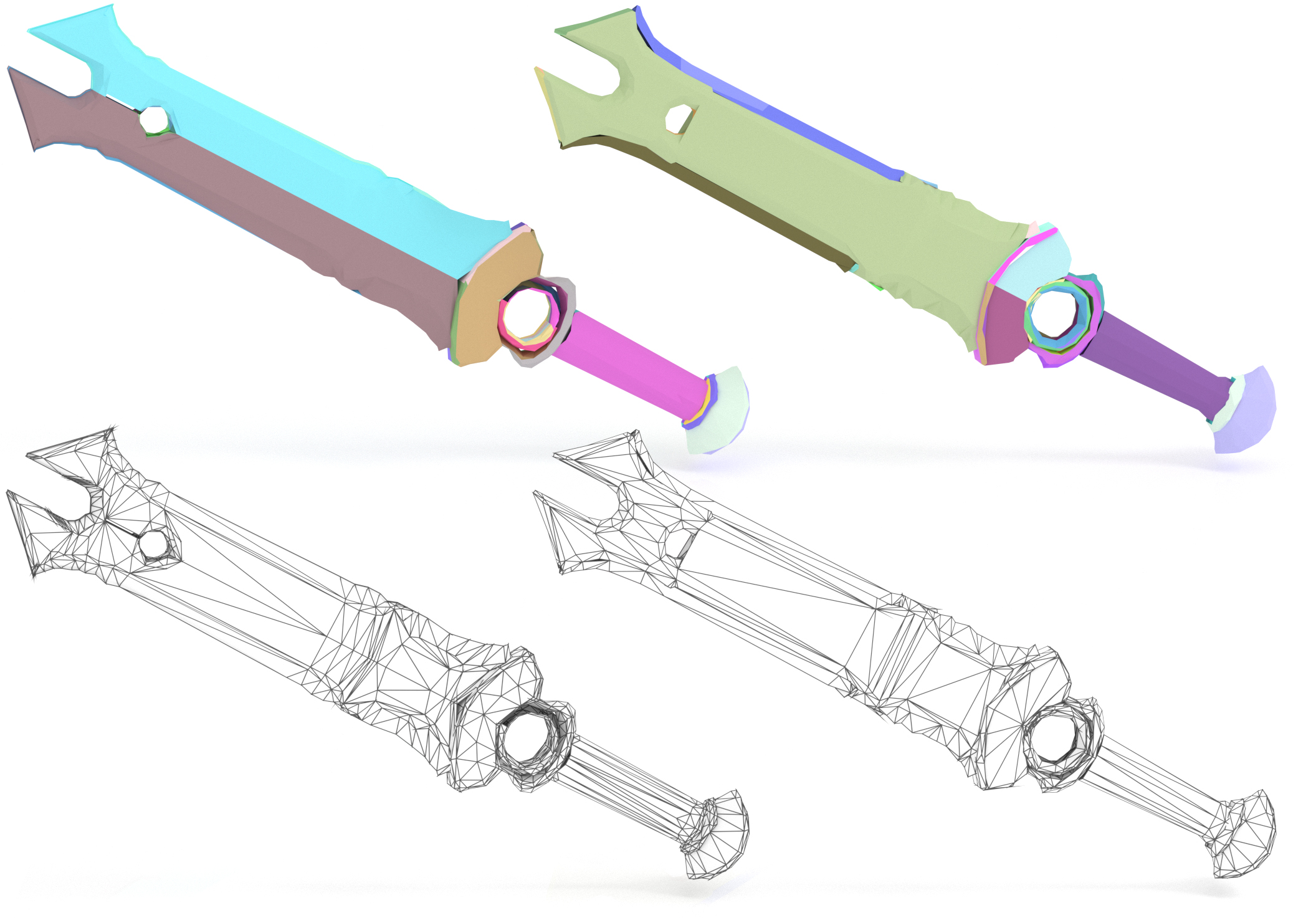}
    \end{overpic}
    \vspace{-8mm}
    \caption{\textbf{Diversity results.}
Conditioned on the same input, our model generates diverse meshes and segmentation outcomes, demonstrating strong generative diversity.}
    \label{fig:diversity}
    \vspace{-4mm}
\end{figure}

\paragraph{Diversity.}
As with other generative models, we showcase the diversity of our outputs in Fig.~\ref{fig:diversity}, which is an important property for all generative systems. Our model produces not only diverse mesh geometries, but also diverse UV segmentation. Despite this diversity, the generated UV charts remain clean, well-structured, and often symmetric, closely reflecting common artist modeling and layout conventions.

\section{Limitations and Future Work}
\label{sec:limit_future}

As the first framework to jointly model mesh generation and UV segmentation while supporting both triangle and quadrilateral outputs, our approach introduces several natural trade-offs.

\begin{figure}[!tp]
    \centering
    \begin{overpic}[width=\linewidth]{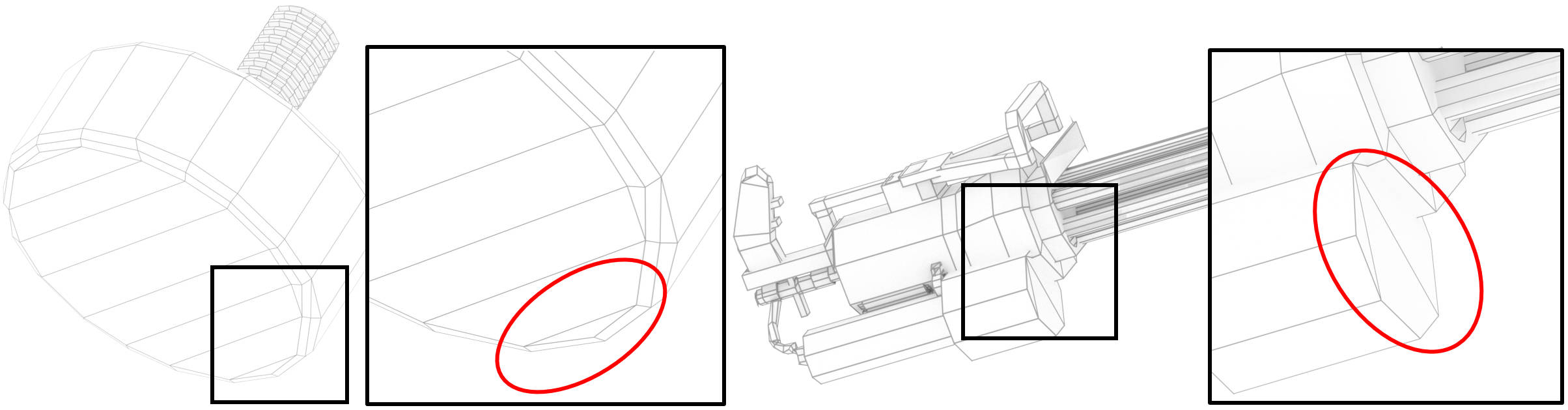}
    \end{overpic}
    \vspace{-8mm}
    \caption{\FFF{\textbf{Failure cases.} Left: Degenerate triangle faces arising from quadrilateral strip decoding under irregular strip configurations. Right: Suboptimal quad layouts limited by the scale and consistency of available high-quality quad-mesh training data.}}
    \label{fig:fail}
    \vspace{-2mm}
\end{figure}

First, our quad output is decoded from quadrilateral strips, which yields predominantly quad-dominant meshes in practice. However, in a small number of cases, e.g., when a strip has an odd length or contains repeated vertices, local faces may degenerate into triangles (\FFF{see Fig.~\ref{fig:fail}, left}). Importantly, these cases are structurally well-defined and can be further reduced with improved dataset quality or lightweight post-processing, which we leave to one of our future works.

Second, the attainable quad quality is currently bounded by the scale and consistency of available high-quality quad-mesh datasets. While our method already produces visually compelling quad layouts, dedicated quad-only approaches such as QuadGPT~\cite{QuadGPT} benefit from being optimized specifically for this setting. We view this as a complementary strength: our goal is a unified model that transfers strong priors from large-scale triangle data and extends them to quad meshes with minimal specialization (\FFF{see Fig.~\ref{fig:fail}, right}).

% First, our quad output is decoded from quadrilateral strips, which yields predominantly quad-dominant meshes in practice. However, in a small number of cases, e.g., when a strip has an odd length or contains repeated vertices, local faces may degenerate into triangles. Importantly, these cases are structurally well-defined and can be further reduced with improved dataset quality or lightweight post-processing, which we leave to one of our future works.

% Second, the attainable quad quality is currently bounded by the scale and consistency of available high-quality quad-mesh datasets. While our method already produces visually compelling quad layouts, dedicated quad-only approaches such as QuadGPT~\cite{QuadGPT} benefit from being optimized specifically for this setting. We view this as a complementary strength: our goal is a unified model that transfers strong priors from large-scale triangle data and extends them to quad meshes with minimal specialization.

Finally, we occasionally observe less regular edge routing on near-spherical shapes (like the bottom left shape in Fig.~\ref{fig:mosaic}). This appears tied to data bias: many triangle datasets represent spheres using near-equilateral tessellations, whereas high-quality spherical exemplars are comparatively scarce in existing quad corpora. Consequently, quad fine-tuning provides a consistent but incremental improvement rather than fully resolving this corner case. We expect this gap to narrow as richer quad datasets become available and as we incorporate stronger shape-adaptive routing priors.

\section{Conclusion}
We present Strips as Tokens (\name), an autoregressive framework for generating high-quality artist meshes with native UV segmentation.
Our strip-based tokenization follows the edge flow of artist meshes and encodes UV island boundaries directly in the sequence, encouraging clean topology and well-structured UV chart partitions.
Building on the same sequence format, \name admits a unified triangle/quad interpretation, enabling mixed-data training that transfers and strengthens priors across formats.
% Extensive experiments show that \name produces diverse, high-fidelity meshes with stronger topological quality than competitive baselines.
Extensive experiments show that \name produces diverse, high-fidelity meshes with stronger topological quality than competitive baselines, highlighting its practical potential for downstream content creation pipelines.

% \section{Conclusion}
% We present Strips as Tokens (\name), an autoregressive framework for generating high-quality artist meshes with native UV segmentation.
% Our strip-based tokenization follows the edge flow of artist meshes and embeds UV island boundaries directly in the sequence, encouraging both clean topology and well-structured UV segmentation.
% Building on the same sequence format, \name admits a unified triangle/quad interpretation, enabling mixed-data training where triangle and quad priors mutually reinforce.
% Extensive experiments show that \name produces diverse, high-fidelity meshes with stronger topological quality than competitive baselines.

% \begin{acks}
% The authors would like to thank the anonymous reviewers for their valuable comments and suggestions. This work is supported by the National Key R\&D Program of China (2021YFB1715900), the National Natural Science Foundation of China (62002190, 62272277, 62072284), and the Natural Science Foundation of Shandong Province (ZR2020MF036, ZR2020MF153). Ningna Wang and Xiaohu Guo were partially supported by National Science Foundation (OAC-2007661).
% \end{acks}

\FloatBarrier
\bibliographystyle{ACM-Reference-Format}
\bibliography{sample-base}

\end{document}